\documentclass[10pt,twocolumn,letterpaper]{article}

\usepackage[pagenumbers]{cvpr}

\usepackage{graphicx}
\usepackage{caption}
\usepackage{subcaption}
\usepackage{float}
\usepackage{longtable}
\usepackage{makecell}
\usepackage{listings}
\usepackage{adjustbox}
\usepackage{xcolor}
\usepackage{amsmath}
\usepackage{listings}
\usepackage{svg}

\definecolor{cvprblue}{rgb}{0.21,0.49,0.74}
\usepackage[pagebackref,breaklinks,colorlinks,allcolors=cvprblue]{hyperref}
\def\confName{ICCV 2025 Findings Workshop}
\def\confYear{2025}

\title{LEMUR Neural Network Dataset: Towards Seamless AutoML}

\author{
Arash Torabi Goodarzi, Roman Kochnev, Waleed Khalid, Hojjat Torabi Goudarzi,\\
Furui Qin, Tolgay Atinc Uzun, Yashkumar Sanjaybhai Dhameliya, Yash Kanubhai Kathiriya,\\
Zofia Antonina Bentyn, Dmitry Ignatov, Radu Timofte\\
Computer Vision Lab, CAIDAS, University of Würzburg, Germany\\
\tt\small
\{arash.torabi\_goodarzi, roman.kochnev, waleed.khalid, furui.qin,\\
\tt\small tolgay-atinc.uzun, yashkumar-sanjaybhai.dhameliya, yash-kanubhai.kathiriya,\\
\tt\small zofia.bentyn\}@stud-mail.uni-wuerzburg.de, \\
\tt\small torabigh@oregonstate.edu
}

\begin{document}
\maketitle

\begin{abstract}
Neural networks have become the backbone of modern AI, yet designing, evaluating, and comparing them remains labor-intensive. While many datasets exist for training models, there are few standardized collections of the models themselves. We present LEMUR, an open-source dataset and framework that brings together a large collection of PyTorch-based neural networks across tasks such as classification, segmentation, detection, and natural language processing. Each model follows a common template, with configurations and results logged in a structured database to ensure consistency and reproducibility. LEMUR integrates Optuna for automated hyperparameter optimization, provides statistical analysis and visualization tools, and exposes an API for seamless access to performance data. The framework also supports extensibility, enabling researchers to add new models, datasets, or metrics without breaking compatibility. By standardizing implementations and unifying evaluation, LEMUR aims to accelerate AutoML research, facilitate fair benchmarking, and lower the barrier to large-scale neural network experimentation. To encourage adoption and collaboration, LEMUR and its plugins are released as open-source projects under the MIT license and are available at: 
\href{https://github.com/ABrain-One/nn-dataset}{https://github.com/ABrain-One/nn-dataset}, 
\href{https://github.com/ABrain-One/nn-plots}{https://github.com/ABrain-One/nn-plots}, and 
\href{https://github.com/ABrain-One/nn-vr}{https://github.com/ABrain-One/nn-vr}.

\end{abstract}

\section{Introduction}
\label{sec:intro}

The rapid progress of artificial intelligence is driven by increasingly powerful neural network architectures. Yet, designing, tuning, and benchmarking these models is still a slow and resource-heavy process that depends on extensive manual effort. Traditional datasets provide raw inputs such as images or text, but researchers lack standardized datasets of the models themselves—datasets that could make evaluation and automation far more efficient.

This paper introduces the neural network dataset LEMUR -- an open-source collection of rigorously tested neural network implementations, which is accompanied by a set of tools in form of a framework for data management, hyperparameter optimization, and model evaluation. LEMUR, which stands for Learning, Evaluation, and Modeling for Unified Research, is designed to provide a unified structure for diverse neural network architectures. Unlike existing projects and repositories such as Torchvision \cite{PyTorch}, Hugging Face \cite{huggingface} and other similar projects and repositories \cite{crossedwires, younger, modelzoos, dl-models, wolfram-models, ONNXModelZoo}, our dataset ensures standardized model implementations, enforcing a cohesive format across all included architectures. Each model adheres to a common template, facilitating seamless evaluation, hyperparameter tuning, and cross-comparison.

A unique key innovation of LEMUR is its maintainability and extensibility. The framework allows users to effortlessly contribute new architectures while ensuring consistency in implementation and evaluation. All models are stored in a structured format, with configurations and results systematically logged in an SQLite database, ensuring efficient data management and retrieval.

Another core feature is its automated testing and evaluation framework, which streamlines performance analysis through reproducible experiments. The dataset is fully integrated with Optuna for hyperparameter optimization, enabling systematic performance tuning across different tasks. Additionally, it incorporates a robust statistical analysis toolkit, allowing researchers to generate graphical insights, compare architectures, and analyze training dynamics automatically.

With built-in support for image classification, segmentation, object detection, and natural language processing, LEMUR is a comprehensive resource for research and practical applications. It excels in AutoML by streamlining model selection and hyperparameter tuning while also supporting edge devices for efficient deployment in resource-constrained environments. Large language models, driven by their growing complexity and scale, are increasingly replacing traditional methods across various domains \cite{Gado2025llm, Rupani2025llm}, including those focused on the automated exploration of extensive hyperparameter spaces. 


The remainder of this paper explores related work in \cref{sec:works}, details our contributions in \cref{contribution}, and presents an overview of the proposed framework in \cref{sec:structure}. We discuss the deep learning tasks supported in \cref{sec:dl-tasks}, followed by a breakdown of the statistical tools and the accompanying performance evaluation framework in \cref{sec:usage}. Finally, we conclude with future research directions in \cref{sec:future}.

\subsection{Related Work}
\label{sec:works}
In the domain of neural networks, datasets have always played a central role in enabling training, evaluation, and benchmarking. Large-scale labeled datasets such as ImageNet for vision tasks \cite{ImageNet} and GLUE for natural language processing \cite{GLUE} have been the driving force behind many breakthroughs in deep learning. Over the years, a wide variety of resources have been developed for images, text, and even biological data \cite{maas2011learning, krizhevsky2009learning, lin2015microsoftcococommonobjects, sweeney2018unsupervised, uniprot}. These collections provide the raw inputs needed for learning representations. However, what has been missing is a curated dataset that captures the neural network architectures themselves—their structural variations, hyperparameters, and performance behaviors—rather than just the raw data they consume.

The motivation for building such a dataset grew out of the limitations of existing model libraries. For example, the PyTorch Vision Library \cite{PyTorch} provides access to many popular pretrained models. Yet these models are not organized into a unified, standardized dataset that records architecture details, hyperparameters, and benchmarking statistics. Similarly, Hugging Face Transformers \cite{huggingface} has become a dominant hub for pretrained models in NLP and beyond, but, like PyTorch Vision, it does not enforce normalization or standardization that would allow seamless cross-domain evaluation or consistent workflows. In both cases, the models exist, but the infrastructure to treat them as part of a structured dataset is absent.

Another important context is the rise of AutoML frameworks, which aim to automate the optimization and evaluation of machine learning models. These frameworks have shown how standardized benchmarking and workflow normalization can accelerate progress and improve reproducibility \cite{Hutter2019, McLaughlin2015}. In parallel, advances in transfer learning \cite{Zoph2018} and novel loss functions \cite{Lin2018} demonstrate the importance of having well-defined evaluation pipelines that make it easy to adapt, extend, and compare models. Despite this progress, no existing framework integrates these principles into a cohesive dataset of neural networks themselves. Current approaches automate training and tuning but do not address the lack of a structured, maintainable collection of models and their associated metadata.

Several attempts have been made to build repositories of neural networks for specific purposes. The CrossedWires dataset \cite{crossedwires} highlights cross-framework reproducibility issues by showing that identical models implemented in different frameworks can produce significantly different accuracies. The Younger dataset \cite{younger} takes a different approach by compiling more than 7,600 unique AI-generated architectures from a pool of over 174,000 real-world models across 30 tasks. This dataset serves as a benchmark for graph neural networks and as a resource for studying architecture diversity. Other repositories fill educational or practical niches, such as Raschka’s deeplearning-models collection \cite{dl-models}, which provides TensorFlow and PyTorch implementations mainly for teaching, or the Wolfram Neural Net Repository \cite{wolfram-models}, which hosts pretrained models for immediate use. The ONNX Model Zoo \cite{ONNXModelZoo} also contributes by enabling cross-framework compatibility with standardized ONNX models. Collectively, these resources enrich the ecosystem, but each is limited in scope: some focus on pretrained weights, others on architecture diversity, and others on framework compatibility.

The LEMUR dataset is designed to address this gap. It provides an expandable and standardized repository of PyTorch-based neural network implementations, each following a consistent template. Unlike ad-hoc collections, LEMUR couples these implementations with a unified evaluation and benchmarking framework, integrating hyperparameter tuning, structured logging, and automated statistical analysis. By treating models themselves as the “data,” LEMUR automates performance verification across training setups and data transformations, producing consistent performance statistics for each model. This design ensures that model evaluations are reproducible, comparable, and extensible.

We argue that LEMUR’s explicit focus on standardization, reproducibility, and unified evaluation positions it as a valuable addition to the deep learning ecosystem. It complements existing datasets and repositories by providing the missing layer of structured model-level benchmarking. This enables researchers not only to access diverse architectures but also to analyze them under a consistent framework, which is essential for progress in areas such as AutoML, model comparison, and reproducible research.

\begin{figure*}[!h]
    \centering
    \includegraphics[width=0.95\textwidth]{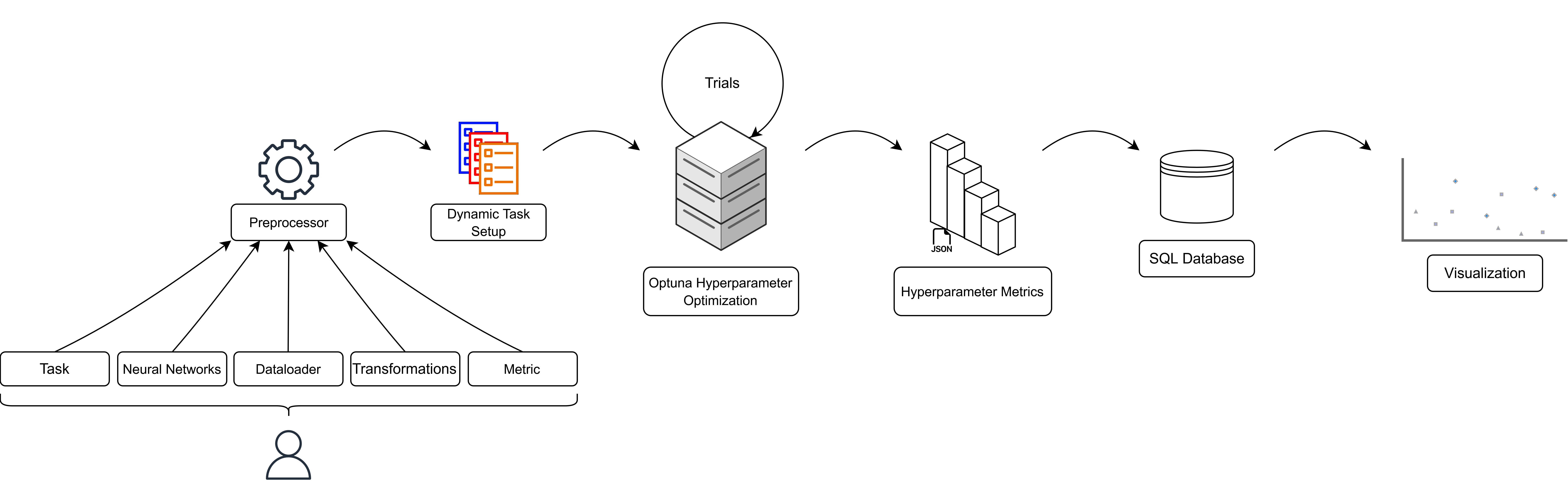}
    \caption{\label{fig:hand}A high level illustration of the LEMUR pipeline, including components like preprocessing, dynamic task allocation, JSON handling, Optuna hyperparameter optimization with additional user specified neural networks, data loaders, transformations, and metric evaluation.}
    \label{fig:pipeline}
\end{figure*}

\subsection{Our Contribution}
\label{contribution}
LEMUR provides a unified infrastructure for benchmarking and extending PyTorch-based neural network models through a combination of standardized implementation templates, schema-driven configuration, and reproducible evaluation workflows. It supports a growing collection of diverse architectures for tasks such as classification and segmentation, each paired with YAML-based specs, automated hyperparameter tuning via Optuna, and detailed accuracy traces stored in an SQLite-backed database. All models are validated for correctness and convergence, with training statistics visualized to enable comparative analysis. By integrating these elements, LEMUR enables reliable experimentation, modular reuse, and seamless integration into AutoML frameworks.

\section{LEMUR Overview}
\label{sec:structure}

\begin{figure*}[!h]
    \centering
    \includegraphics[width=0.95\textwidth]{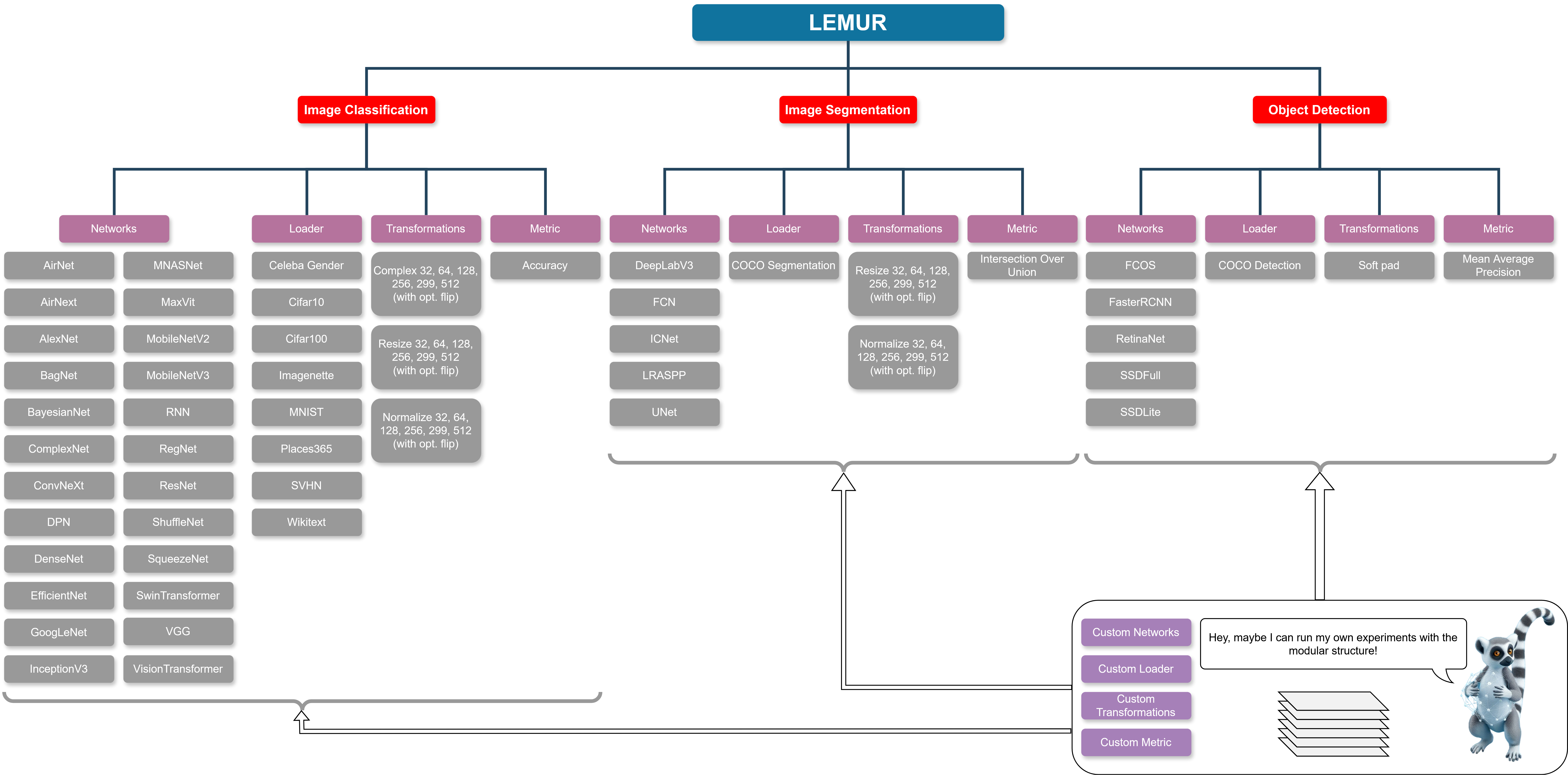}
    \caption{Tree-like diagram illustrating examples of supported tasks, neural network architectures, transformations, metrics, and datasets. The framework's modularity allows users to add custom loaders, networks, and metrics for experimentation.}
    \label{fig:supported}
\end{figure*}

In this section, we discuss the structure of the neural network dataset and its supporting framework. The dataset is continuously expanding and currently includes 136 model implementations covering 52 unique architectures. The proposed LEMUR framework offers a comprehensive set of features designed to support reproducibility, extensibility, and analysis.

A central capability of LEMUR is its experiment tracking system. For every run, the framework automatically records all relevant configurations, including hyperparameters, dataset identifiers, and evaluation metrics, as detailed in \cref{optuna}. These records are stored in both JSON files and a SQL database, ensuring that no experiment is lost and that results can be easily retrieved or extended. This mechanism enables detailed analysis and guarantees reproducibility, which is critical for advancing neural network research.

To support interpretability, the framework integrates results into statistical summaries and automatically generates visualizations of key trends. In addition, all summaries can be exported in Excel format, making performance reports accessible to users with different technical backgrounds. This reduces the need for manual data handling and allows researchers to move directly from training to analysis and interpretation.

Flexibility is another defining principle of LEMUR. The framework supports the seamless integration of additional scripts, datasets, or metrics, making it suitable for a wide range of research scenarios. Its modular design ensures that both novice and advanced users can benefit: newcomers can run experiments and analyze results with minimal setup, while experienced researchers can extend the system to meet domain-specific needs.

The framework also streamlines data management. Unique identifiers are assigned to every record to prevent duplication, and all configurations, metrics, and results are logged automatically during training. By combining JSON storage for transparency with a database backend for structured queries, the framework creates a reliable foundation for large-scale benchmarking and reproducible research.

The general pipeline is depicted in \cref{fig:pipeline}. To launch an experiment, users specify the task, dataset loader, preprocessing transformations, evaluation metric, and neural network architecture, as illustrated in \cref{listing:configuration_code}.

\begin{lstlisting}[language=bash, caption=Code Block for Configuration, label={listing:configuration_code}]
python run.py -c <task-name><dataset-name><metric>_<neural-net-name>
\end{lstlisting}

Once initialized, the pipeline sets up an Optuna optimization task dynamically according to the configuration. A series of trials is then executed, with each trial exploring different hyperparameter combinations. Results from each trial are first saved as JSON files and then transferred into the SQL database, from which graphical insights and comparative analyses are generated automatically.

This pipeline design ensures flexibility, efficiency, and extensibility. By storing results in a structured database, the framework allows seamless querying, comparison, and retrieval of performance metrics across models and tasks. These records can be used to generate statistical summaries, performance trend analyses, and comparative visualizations, providing deeper insights into model behavior. Moreover, the modular nature of the pipeline allows users to extend the framework with new architectures, tasks, transformations, and metrics without modifying the core system.

We ran these experiments for all the supported configurations by default and included their results in the LEMUR package.
These results can be accessed in form of a Pandas Dataframe \cite{pandas} via the included API.
This can be done by calling the function {data()} from the API which is available just by importing the library.
This function returns a dataframe object with the following columns;
"task", "dataset", "metric", "metric\_code", "nn", "nn\_code", "epoch", "accuracy", "duration", "prm" and "transform\_code".
This function also allows flexible filtering of the returned data, enabling users to retrieve only the best achieved accuracy for each model, filter models based on their designated tasks, include specific types of model architectures, and apply additional filters by setting the appropriate parameters when calling the function.
Additionally, it allows users to generate a variety of plots and output in Excel format that provide insights into the performance of neural network models, as discussed in \cref{visualization}, facilitating a comprehensive exploration of experimental outcomes.

This API ensures the ease of access to preprocessed data and ensures that the LEMUR framework is applicable for integration a numerous kinds of projects. Furthermore, by leveraging this structured API, users can efficiently analyze model performance trends across different architectures and datasets without the need for manual data handling. This streamlined access to experimental results enhances reproducibility and allows researchers to seamlessly incorporate LEMUR’s insights into their own machine learning workflows.

For a comprehensive explanation of the design of the accompanying SQL database, please refer to the supplementary material. However, we will outline the framework's directory structure and unified scheme for neural network implementation in the remainder of this section, as it is fundamental to extend the framework's scope and incorporating it in the different workflows.

\subsection{Directory Structure}
\label{directories}
The entire project, together with its dependencies, is gathered in form of a python package named {"ab"}.
The main part of the project, which includes the dataset itself, the API and the most important and commonly used tools provided with the framework, is a sub-package of {"ab"} named {"nn"}.
Other extensions such as advanced statistical tools (see \cref{visualization}) for the project can be installed and will be placed next to {"nn"}.

The package {"nn"} is itself is subdivided into several directories. 
The neural network implementations provided in the NN-Dataset framework are placed in a directory named {"nn"} ({ab.nn.nn}). Each of the neural network models included, is implemented in its distinctive Python file named after its model structure. Each model is assigned a unique name, and when new models are added through the framework's API ({ab.nn.api}), the system automatically generates unique identifiers for them, if not provided. This mechanism eliminates potential naming conflicts and facilitates the seamless integration of new models.

Our framework provides different performance metrics and data transformation techniques, which are suitable for the deep learning tasks supported. These methods are implemented in {ab.nn.metric} and {ab.nn.transform} respectively.

The experimentation results done by our team are provided in {ab.nn.stat} which includes a directory for each of the testing configurations we have tested. For more information about configurations, refer to \cref{sec:dl-tasks}.
The details of this experimentation can be followed in {ab.nn.train}, through which the model evaluation capabilities of our framework are facilitated.

\subsection{Unified Code of Neural Networks}
\label{code-structure}
To achieve the goal of unifying neural network implementations across various tasks, a standardized code structure is essential. This structure must support flexible training loops, adaptable training and testing datasets, and customizable hyperparameters for different models while maintaining a consistent interface. Such a design ensures that all included neural networks can be evaluated and analyzed in a cohesive and uniform way.

Each neural network in the dataset is implemented as a unique Python file containing a PyTorch \texttt{nn.Module} class that adheres to the following standards:

\begin{enumerate}
    \item \textbf{Main Class Structure:}  
    The primary class in each file is named \texttt{Net}. It must be initialized with three arguments: the first argument specifies the input shape of the model as a tuple of integers, and the second argument specifies the output shape in the same format. The third argument is a dictionary named \texttt{prm} that contains additional parameters specific to the model, including hyperparameters for the model structure and its training process.
    
    \item \textbf{Learning Loop:}  
    Each model must implement a method named \texttt{learn(train\_data)}, which accepts iterable training data as input and updates the model parameters based on the defined learning loop.
    
    \item \textbf{Training Setup:}  
    A method \texttt{train\_setup(device, prm)} must initialize training by setting up the optimizer, defining the loss function from \texttt{prm}, and selecting the computational device (e.g., CPU, GPU or MPS).
    
    \item \textbf{Supported Hyperparameters:}  
    Each file must include a function named \texttt{supported\_hyperparameters()} that returns a set of hyperparameters supported by the model’s structure and training process. These hyperparameters are expected to be provided in \texttt{prm}.
\end{enumerate}

This standardized structure ensures consistency and interoperability, allowing for seamless evaluation and analysis of all included neural networks. It furthermore allows for flexibility in the training loop for specific neural network structures while maintaining a cohesive framework for benchmarking and comparison. The design ensures that hyperparameters remain flexible, allowing support for any neural network architecture without imposing rigid constraints. This adaptability enables seamless integration of diverse models while preserving a consistent evaluation pipeline. Additionally, the structured approach facilitates automated hyperparameter tuning, performance evaluation, and dataset adaptability, making the framework highly scalable and efficient for both research and practical applications.

\section{Supported Deep Learning Use Cases}
\label{sec:dl-tasks}
To accommodate task-specific training and testing datasets, we provide a range of dataset loaders accessible through a provided parser. This approach ensures flexibility to expand supported datasets for various deep learning tasks while maintaining a unified operational interface for consistent high-level training.

In this section, we discuss the most important deep learning tasks currently supported by the LEMUR framework.
For each of these tasks, we explain the supported data loading and preprocessing techniques, as well as the performance metrics provided for model evaluation.

The LEMUR framework is designed to support three fundamental deep learning tasks by default: \textbf{image classification, image segmentation, and object detection}. Each of these tasks is equipped with dedicated data loaders, preprocessing transformations, evaluation metrics, and hyperparameter configurations, ensuring a standardized and reproducible workflow across different architectures. The supported configurations for each of these three tasks is depicted in \cref{fig:supported}. These result in 9,158 unique configurations. As a demonstration, we have included the accuracy results of the included models in the supplementary material.

For \textbf{image classification}, the framework provides loaders for widely used datasets such as MNIST, CIFAR-10, and CIFAR-100, allowing for easy benchmarking of model performance. The preprocessing pipeline includes essential transformations such as normalization and resizing, while accuracy serves as the primary evaluation metric. Users can experiment with various architectures, tuning hyperparameters like learning rate, batch size, and optimizer settings to assess model behavior across datasets.

\textbf{Image segmentation} is supported through advanced data handling techniques that enable efficient mask generation and transformation. The framework provides built-in support for segmentation datasets, including COCO-Seg 2017, and incorporates preprocessing steps such as category reduction and custom mask filtering. The primary performance metric for segmentation tasks is mean Intersection over Union (IoU), ensuring that models are evaluated based on the accuracy of predicted segmentation masks. Additionally, the framework optimizes data augmentation strategies to enhance model generalization.

For \textbf{object detection}, the framework integrates loaders that handle bounding box annotations and ensure efficient batch processing. The dataset structure follows best practices for detection tasks, including automatic category reduction and adaptive transformations to preserve annotation accuracy. The mean Average Precision (mAP) metric is used to evaluate detection performance, providing a standardized approach for comparing object localization across different models. To streamline experimentation, the framework supports hyperparameter tuning for detection-specific parameters such as anchor box sizes and IoU thresholds.

Each of these tasks is designed with flexibility and extensibility, allowing researchers to test various neural network architectures while maintaining a unified implementation framework. Additionally, the integration of automated hyperparameter tuning and statistical performance analysis enables systematic model comparison across tasks.

For a more detailed breakdown of dataset loaders, preprocessing techniques, and hyperparameter configurations for each deep learning task, please refer to the supplementary material.

\section{Usage}
\label{sec:usage}
In this section, we discuss the design of the most prominent analysis tools of our framework, namely, the neural network performance analysis with Optuna in \cref{optuna} and the tools for statistical analysis and data visualization in \cref{visualization}.

\subsection{Performance Exploration with Optuna}
\label{optuna}

Optimizing model performance by selecting effective hyperparameters is a persistent challenge. Traditional methods like grid or random search~\cite{JMLR:v13:bergstra12a} are inefficient and computationally expensive. To address this, we incorporate Optuna~\cite{optuna}, a modern hyperparameter optimization framework, into LEMUR for large-scale performance tuning.

Optuna employs the Tree-structured Parzen Estimator (TPE)~\cite{NIPS2011_86e8f7ab}, a Bayesian approach that adaptively focuses on promising regions of the hyperparameter space. This improves convergence and reduces compute costs. Its lightweight design integrates easily into modern ML workflows.

A key strength is its ability to handle complex, high-dimensional search spaces. Users can define arbitrary hyperparameter combinations tailored to their models. Optuna then automates the trial execution, testing parameters like learning rate, batch size, and momentum across diverse architectures.

Each trial is logged with detailed metrics—training accuracy, validation loss, and epoch-wise trends, allowing for thorough performance analysis. By default, 41,131 Optuna trials were conducted to match the size of the dataset.

This setup supports reproducible benchmarking and accelerates architecture-level comparisons. Optuna’s flexibility and automation make it well suited for scalable experimentation and integration with AutoML frameworks.

Full details on experimental setups and optimization configurations are provided in the supplementary material.

\begin{figure}[H]
    \centering
    \includegraphics[width=0.45\textwidth]{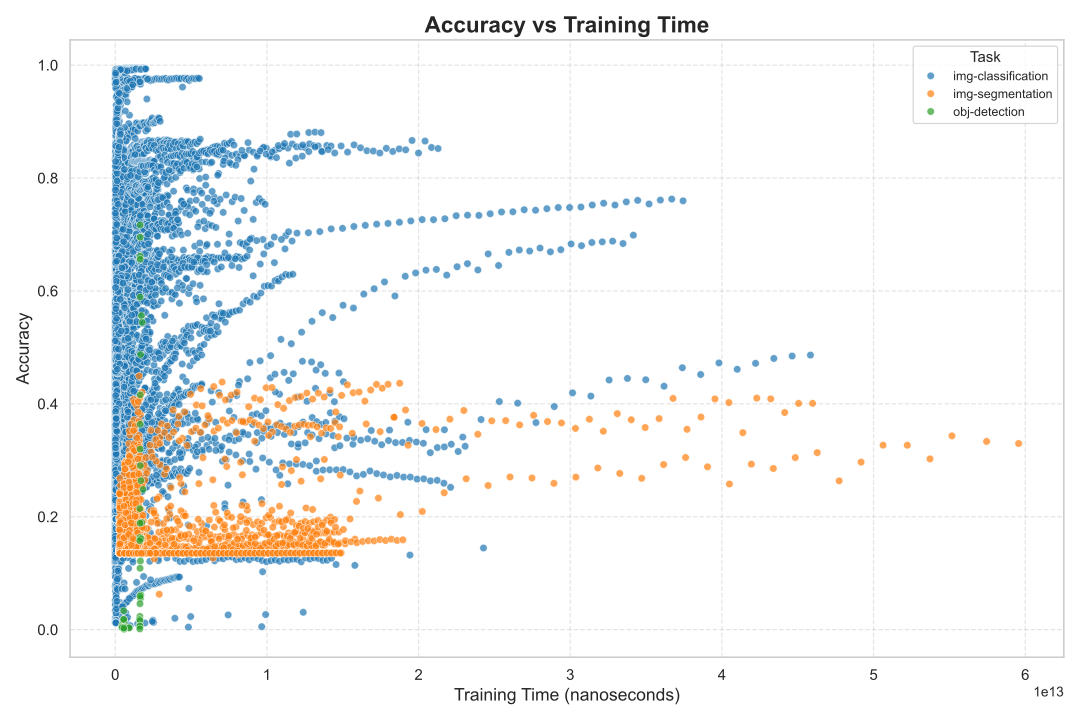}
    \caption{Scatter plot showing the relationship between accuracy and training time (in nanoseconds) for different tasks (image classification, image segmentation, and object detection). Image classification demonstrates rapid accuracy improvement with lower training times and achieves high accuracy consistently. Image segmentation exhibits slower improvements with moderate accuracy, while object detection has lower initial accuracy and requires longer training times to stabilize. The plot emphasizes the varying computational demands and learning behaviors across tasks}
    \label{fig:scatter_duration}
\end{figure}

\begin{figure}[H]
    \centering
    \includegraphics[width=0.45\textwidth]{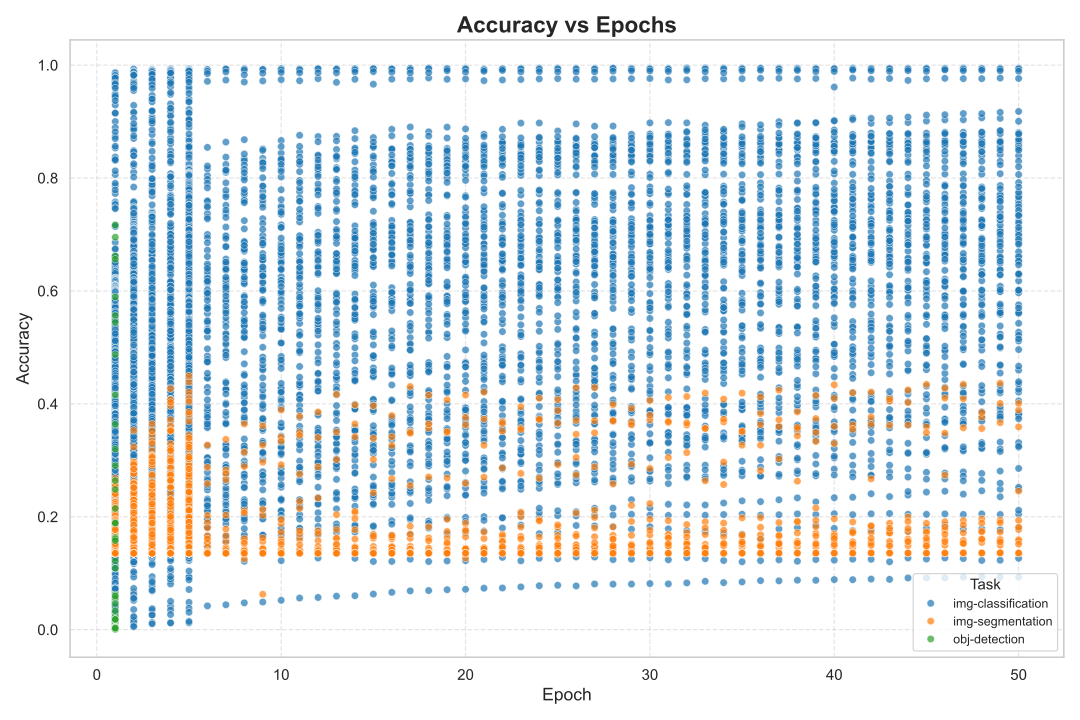}
    \caption{Scatter plot illustrating the variation of accuracy across epochs for different tasks (image classification, image segmentation, and object detection). The plot highlights distinct task-specific accuracy trends, with image classification showing faster and more consistent improvements, while image segmentation and object detection exhibit lower initial accuracy and more gradual improvements. Consistency in accuracy increases with epochs for all tasks, reflecting convergence and stabilization over time.}
    \label{fig:scatter_epochs}
\end{figure}

\subsection{Statistics \& Data Visualization}
\label{visualization}
Statistical data visualization serves as a cornerstone for understanding and interpreting the complex performance metrics associated with neural networks. In the proposed LEMUR framework, visualization is instrumental in uncovering trends, identifying patterns, and evaluating model performance across a diverse range of tasks and datasets. 

The LEMUR framework employs a comprehensive and well engineered set of visualization techniques to analyze and interpret both raw and aggregated data. These visualizations uncover trends, variability, and performance dynamics across different tasks and datasets. The combination of scatter, line, box, histogram, rolling mean, mean/standard deviation plots, and duration distribution visualizations provides a comprehensive view of model performance across tasks and datasets.
Below, we describe the insights drawn from a selection of these plots. 

For a comprehensive description of the workflow used to generate the visualizations, as well as additional plots and insights not included in the main paper, please refer to the supplementary material.

\paragraph{Raw Data Visualizations:}
\begin{figure*}[ht]
    \centering
    \begin{minipage}{0.33\linewidth}
        \centering
        \textit{(a) Rapid convergence with stabilization above 0.8 accuracy.} \\
        \includegraphics[width=\linewidth]{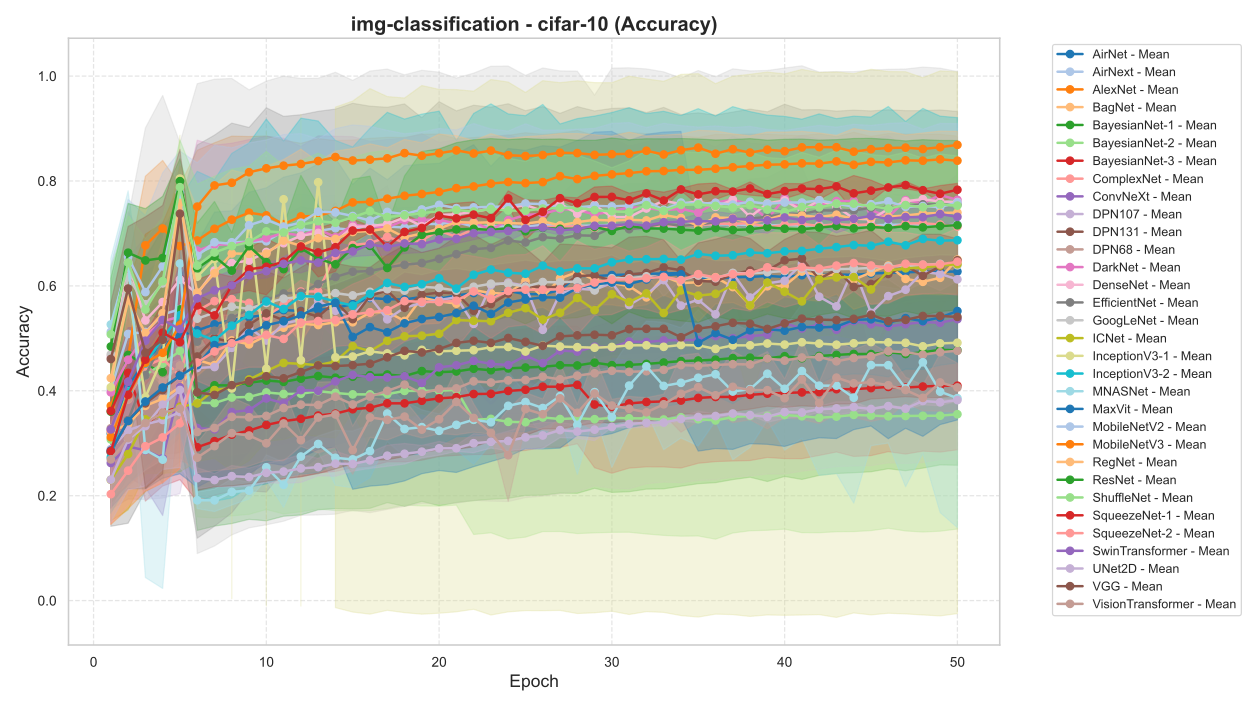}
    \end{minipage}
    \hfill
    \begin{minipage}{0.33\linewidth}
        \centering
        \textit{(b) Gradual improvement, stabilizing below 0.4 accuracy.} \\
        \includegraphics[width=\linewidth]{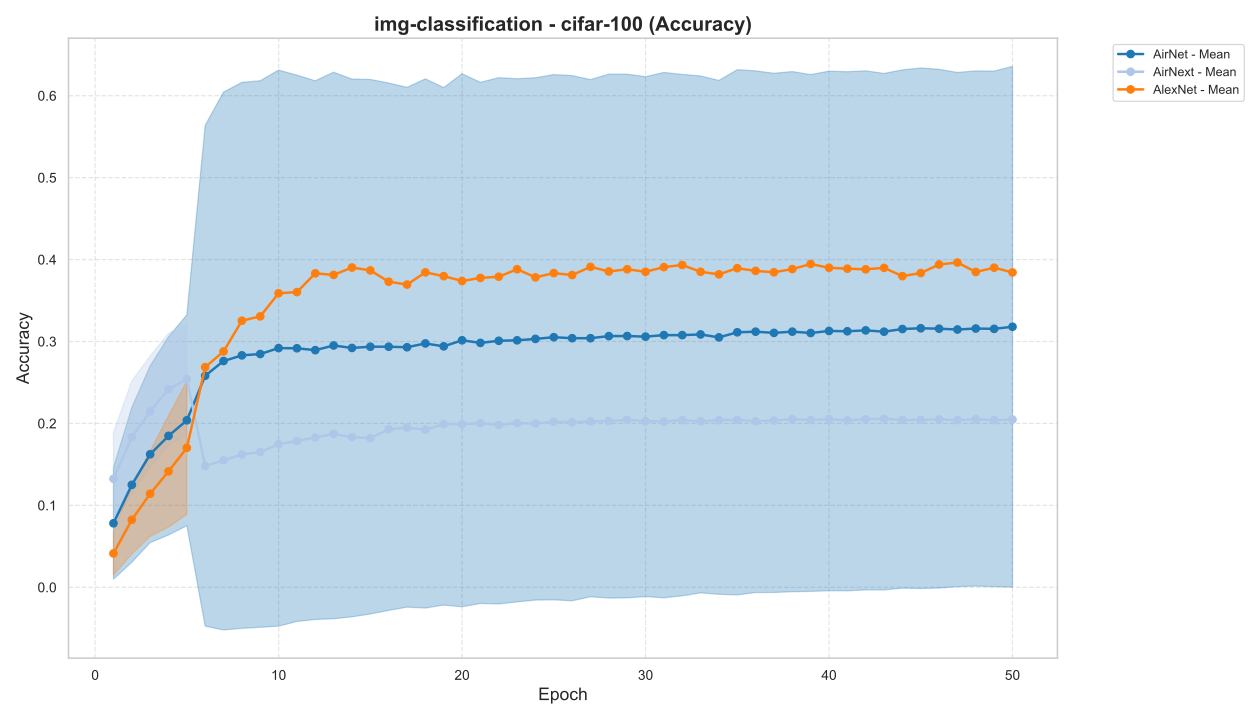}
    \end{minipage}
    \hfill
    \begin{minipage}{0.33\linewidth}
        \centering
        \textit{(c) Quick convergence with variance diminishing in later epochs.} \\
        \includegraphics[width=\linewidth]{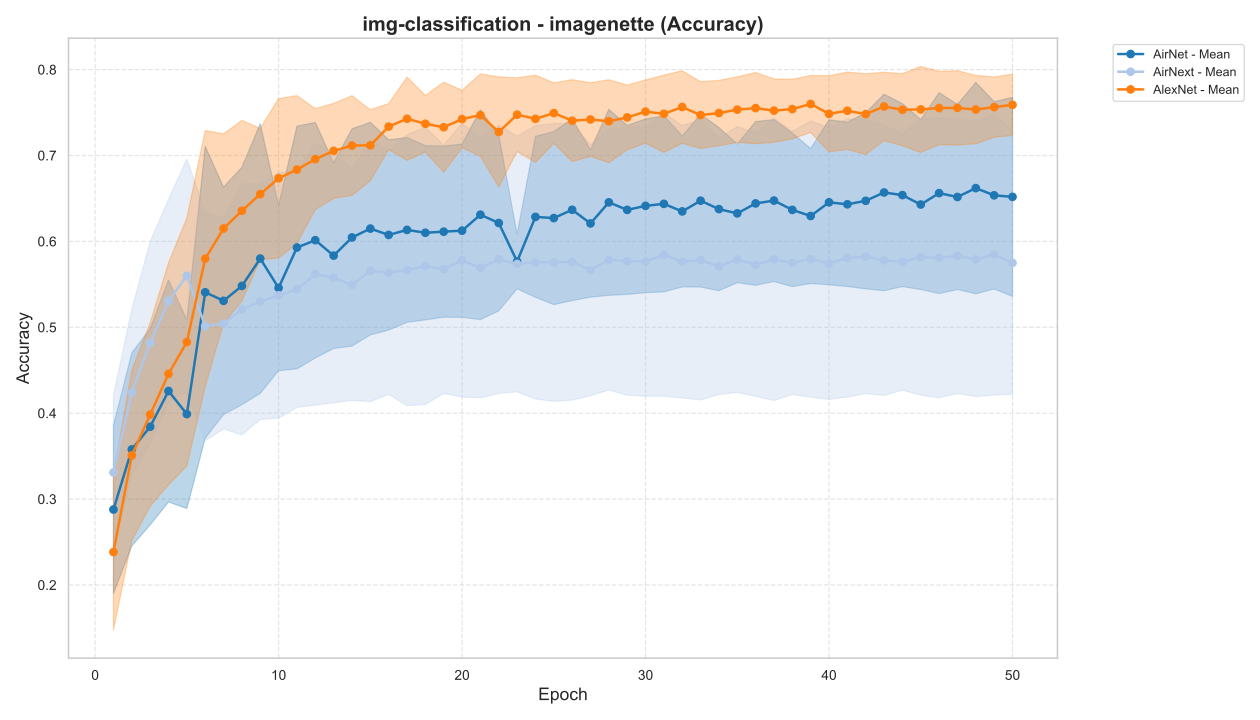}
    \end{minipage}
    \vskip 15pt
    \begin{minipage}{0.33\linewidth}
        \centering
        \textit{(d) Rapid convergence above 0.9 accuracy within 10 epochs.} \\
        \includegraphics[width=\linewidth]{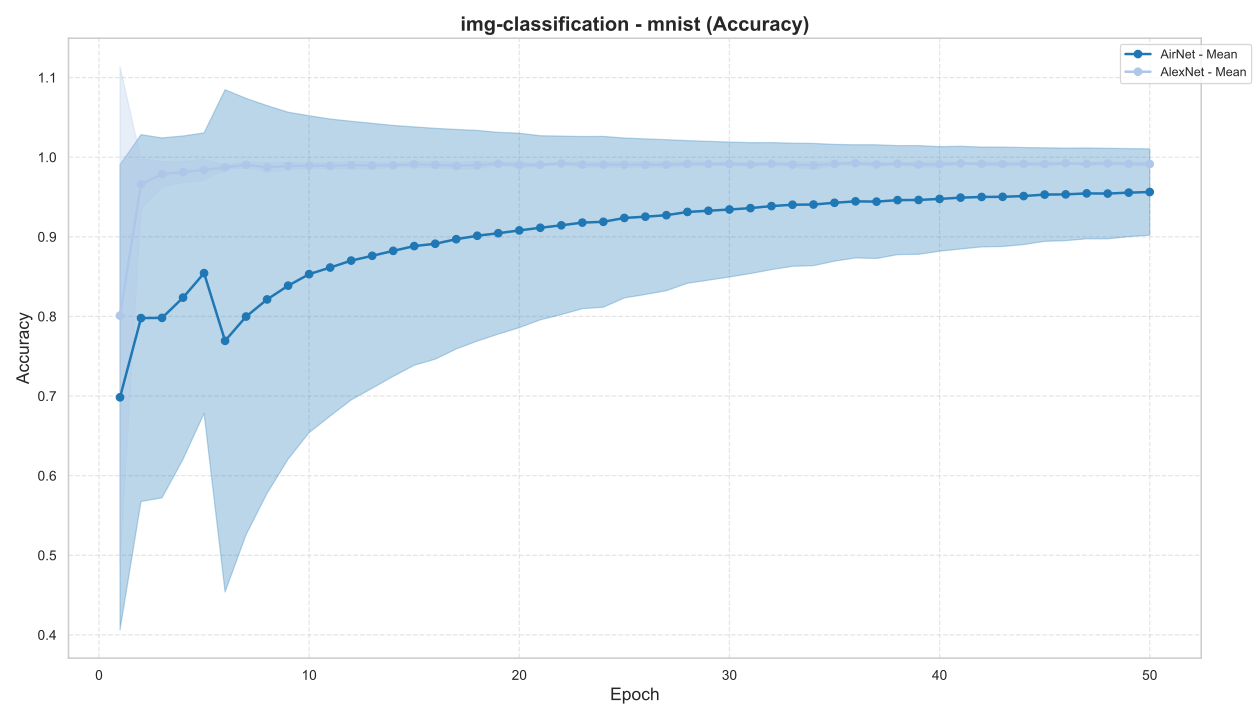}
    \end{minipage}
    \hfill
    \begin{minipage}{0.33\linewidth}
        \centering
        \textit{(e) Slower convergence rates with high initial variance.} \\
        \includegraphics[width=\linewidth]{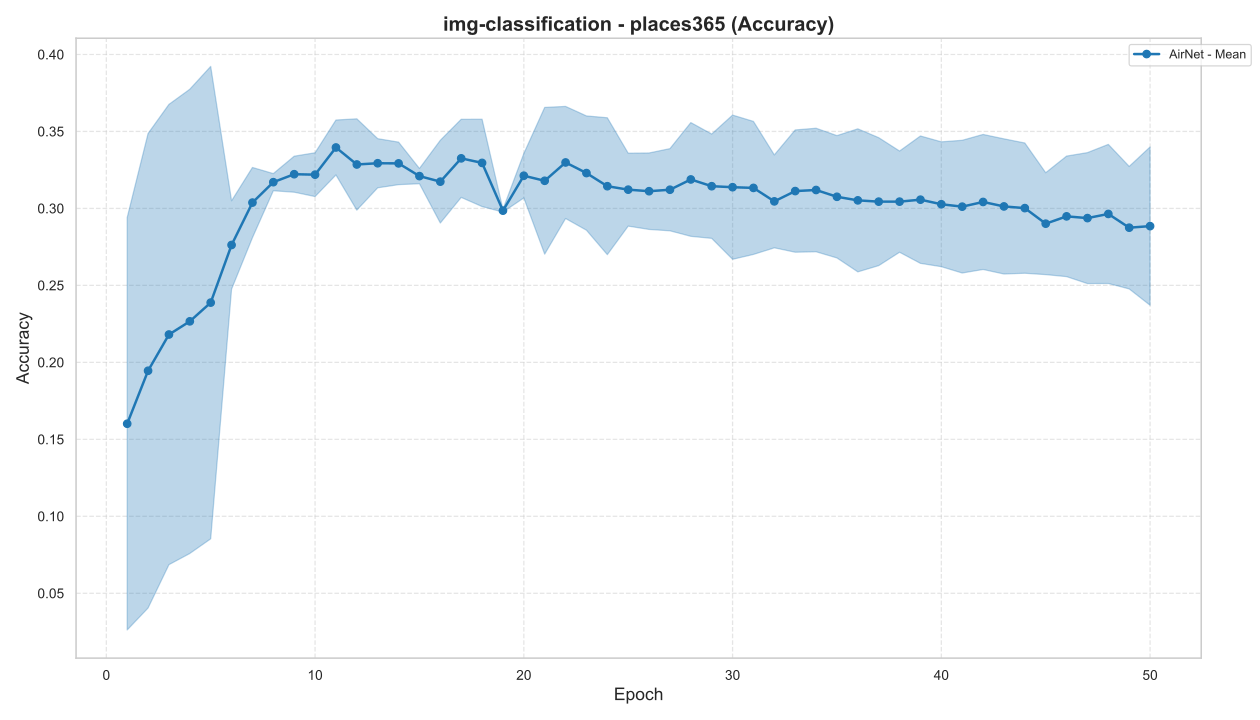}
    \end{minipage}
    \hfill
    \begin{minipage}{0.33\linewidth}
        \centering
        \textit{(f) Steady improvements stabilizing around 0.6-0.8 accuracy.} \\
        \includegraphics[width=\linewidth]{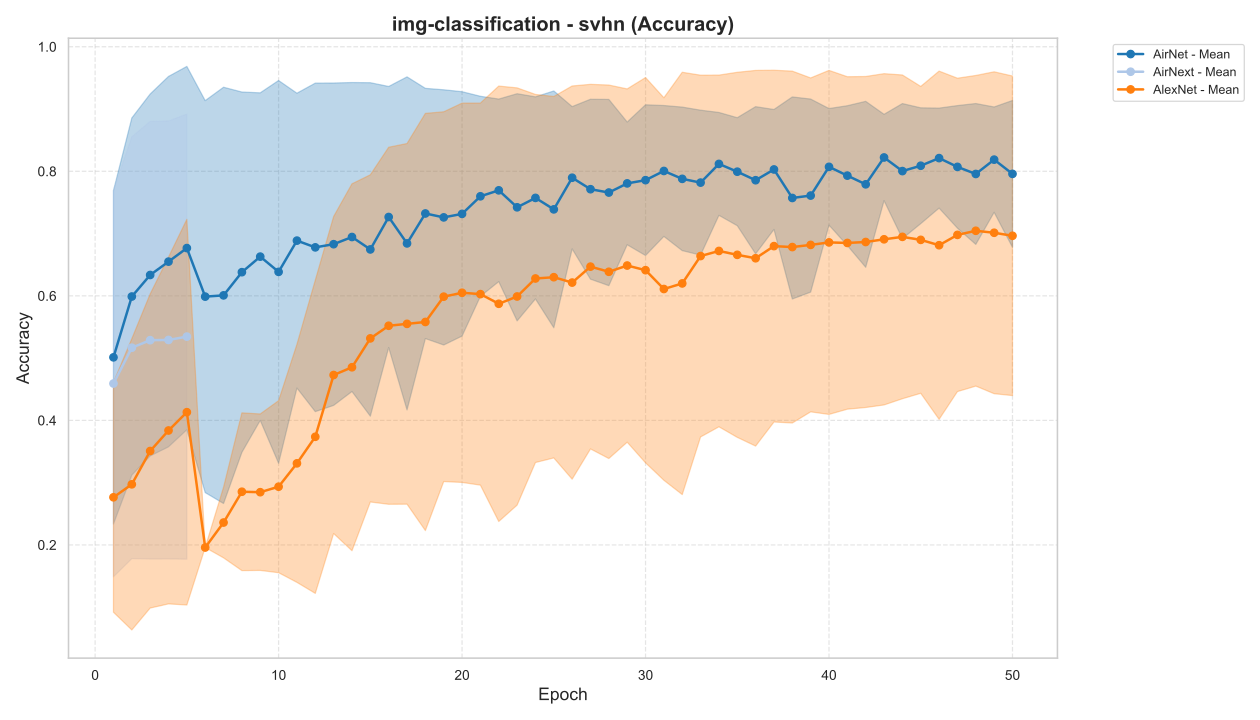}
    \end{minipage}
    \bigskip
    \caption{Accuracy trends across various datasets. Subfigures (a)--(f) highlight the accuracy progression for different models over epochs. Simpler datasets like MNIST and CIFAR-10 show rapid convergence and higher accuracy, while complex datasets like CIFAR-100 and Places365 exhibit slower progress and more variance. Models like AirNet and AlexNet consistently demonstrate robust performance across tasks.}
    \label{fig:accuracy_trends_with_features}
\end{figure*}

Raw data plots provide a granular view of the relationships between training parameters and performance metrics. These visualizations are essential for understanding the initial data characteristics and identifying task-specific patterns and serve to give the first insights in a models training.

The scatter plot \cref{fig:scatter_epochs,fig:scatter_duration} emphasizes the relationship accuracy and training time highlighting task-specific trade-offs and computational demands.

For instance, the scatter plot of accuracy versus epochs (\cref{fig:scatter_epochs}) illustrates task-specific accuracy trends, where image classification exhibits faster and more consistent improvements compared to image segmentation and object detection. Similarly, the scatter plot of accuracy versus training time (\cref{fig:scatter_duration}) highlights the varying computational demands of tasks: image classification achieves high accuracy with lower training times, while segmentation and detection tasks require significantly longer durations.

Furthermore, the LEMUR framework provides box plots, line plots and histograms for raw data visualization to provide statistical insights into metric distributions, showcasing variability and common accuracy ranges achieved by models. Examples of these plots are provided in the supplementary material.

\paragraph{Aggregated Data Visualizations: }
Aggregated visualizations provide a higher-level understanding of model performance by summarizing trends across tasks and datasets. These plots highlight key insights into how training parameters, such as time and epochs, influence performance metrics like accuracy and IoU.

These visualizations include mean and standard deviation plots (\cref{fig:accuracy_trends_with_features}), which capture trends and variability across epochs, providing valuable insights into the consistency of model performance. For example, the subplots in \cref{fig:accuracy_trends_with_features} reveal dataset-specific learning trends: CIFAR-10 (\cref{fig:accuracy_trends_with_features}a) demonstrates rapid convergence and stabilization above 0.8 accuracy, while CIFAR-100 (\cref{fig:accuracy_trends_with_features}b) stabilizes below 0.4 accuracy with more gradual improvements. In contrast, Imagenette (\cref{fig:accuracy_trends_with_features}c) shows quick convergence with diminishing variance over time, and MNIST (\cref{fig:accuracy_trends_with_features}d) achieves over 0.9 accuracy within 10 epochs. Complex datasets like Places365 (\cref{fig:accuracy_trends_with_features}e) exhibit slower convergence with high initial variance, while SVHN (\cref{fig:accuracy_trends_with_features}f) stabilizes between 0.6 and 0.8 accuracy after steady improvements.

\paragraph{Integration of visualizations into Excel reports:}
As a bonus, the LEMUR framework also provides seamless integration of its generated visualizations into comprehensive Excel reports through the \texttt{export\_excel.py} and \texttt{export\_raw\_excel.py} modules. These modules leverage the \texttt{openpyxl} library \cite{openpyxl} to facilitate efficient creation and manipulation of Excel files. The \texttt{export\_excel.py} module embeds aggregated statistical data alongside categorized plots within a structured Excel file, enabling stakeholders to easily interpret and analyze the findings. In contrast, the \texttt{export\_raw\_excel.py} module is specifically designed to export raw data with embedded plots, supporting exploratory analysis and in-depth examination.

By utilizing \texttt{openpyxl}, these modules ensure a high level of customization and compatibility with universally familiar formats. This significantly enhances accessibility, making the reports invaluable for both technical and non-technical audiences. By consolidating visual insights with detailed statistical summaries, the generated reports improve the interpretability and dissemination of findings. This approach aligns with the framework's objective of fostering informed decision-making and ensuring accessibility for diverse stakeholders.

By adhering to this structured workflow, the LEMUR framework ensures the reliability, consistency, and analytical depth of its visual outputs. The integration of automation at each stage of the process minimizes manual intervention, enhances reproducibility, and streamlines the derivation of actionable insights from complex datasets. This approach not only optimizes efficiency but also empowers researchers to explore and interpret data with greater precision.

\section{Conclusion \& Future Work}
\label{sec:future}
In conclusion, this work introduces a comprehensive contribution to the machine learning community through the development of LEMUR, a high-quality dataset of rigorously validated PyTorch neural network implementations, paired with a unified evaluation and analysis framework. The framework combines several novel features: a standardized implementation template that ensures consistency across models, an SQLite-based database for structured experiment tracking and efficient data management, and integrated tools for hyperparameter optimization and statistical analysis \cref{optuna}. Together, these components address persistent challenges in neural network research, particularly those related to reproducibility, comparability, and extensibility. By providing a foundation where models can be benchmarked, fine-tuned, and extended under consistent conditions, LEMUR supports both fundamental research and practical applications in areas such as AutoML, domain adaptation, and optimization studies. 

A defining aspect of the framework is its capacity for growth and adaptability. Built around modular principles, LEMUR allows new architectures, datasets, and task-specific methodologies to be added without breaking compatibility. This ensures that the framework can evolve alongside advances in deep learning, accommodating novel architectures, training strategies, and evaluation metrics as they emerge. The use of a standardized template also guarantees interoperability, meaning that future contributions integrate smoothly into the existing system while preserving reproducibility and consistency across the repository. 

Looking forward, the framework is intended to serve not only as a resource but also as a collaborative platform for the community. By encouraging community contributions, LEMUR can expand into a comprehensive and continuously updated repository that reflects the state of the art in neural network design and experimentation. Such collaboration would broaden its applicability to new domains, support emerging tasks, and enhance its role in advancing reproducibility and standardized evaluation. In particular, extensions into areas such as automated machine learning, transfer learning for domain-specific applications, and advanced optimization strategies could be accelerated through shared development and use, ensuring that LEMUR remains relevant as the field evolves and adaptable to challenges that have yet to emerge.

Ultimately, we envision LEMUR as more than a static dataset or framework—it is designed as a foundation for collective progress. By offering consistency, transparency, and extensibility, it provides researchers and practitioners with the tools to not only conduct reproducible experiments but also to explore innovative approaches in a structured and reliable environment. As the framework grows, it has the potential to become a cornerstone resource for neural network research, lowering barriers for new investigations while supporting advanced studies in optimization, architecture design, and automated machine learning.


{
    \small
    \bibliographystyle{ieeenat_fullname}
    \bibliography{main}

\begin{thebibliography}{34}
\providecommand{\natexlab}[1]{#1}
\providecommand{\url}[1]{\texttt{#1}}
\expandafter\ifx\csname urlstyle\endcsname\relax
  \providecommand{\doi}[1]{doi: #1}\else
  \providecommand{\doi}{doi: \begingroup \urlstyle{rm}\Url}\fi

\bibitem[Aboudeshish et~al.(2025)Aboudeshish, Ignatov, and Timofte]{Aboudeshish2025augmentation}
Nada Aboudeshish, Dmitry Ignatov, and Radu Timofte.
\newblock Augmentgest: Can random data cropping augmentation boost gesture recognition performance?
\newblock \emph{arXiv preprint arXiv:2506.07216}, 2025.

\bibitem[Akiba et~al.(2019)Akiba, Sano, Yanase, Ohta, and Koyama]{optuna}
Takuya Akiba, Shotaro Sano, Toshihiko Yanase, Takeru Ohta, and Masanori Koyama.
\newblock Optuna: A next-generation hyperparameter optimization framework, 2019.

\bibitem[Bergstra and Bengio(2012)]{JMLR:v13:bergstra12a}
James Bergstra and Yoshua Bengio.
\newblock Random search for hyper-parameter optimization.
\newblock \emph{Journal of Machine Learning Research}, 13\penalty0 (10):\penalty0 281--305, 2012.

\bibitem[Bergstra et~al.(2011)Bergstra, Bardenet, Bengio, and K\'{e}gl]{NIPS2011_86e8f7ab}
James Bergstra, R\'{e}mi Bardenet, Yoshua Bengio, and Bal\'{a}zs K\'{e}gl.
\newblock Algorithms for hyper-parameter optimization.
\newblock In \emph{Advances in Neural Information Processing Systems}. Curran Associates, Inc., 2011.

\bibitem[Deng(2012)]{mnist}
Li Deng.
\newblock The mnist database of handwritten digit images for machine learning research [best of the web].
\newblock \emph{IEEE Signal Processing Magazine}, 29\penalty0 (6):\penalty0 141--142, 2012.

\bibitem[Gado et~al.(2025)Gado, Taliee, Memon, Ignatov, and Timofte]{Gado2025llm}
Mohamed Gado, Towhid Taliee, Muhammad~Danish Memon, Dmitry Ignatov, and Radu Timofte.
\newblock Vist-gpt: Ushering in the era of visual storytelling with llms?
\newblock \emph{arXiv preprint arXiv:2504.19267}, 2025.

\bibitem[Gazoni and contributors(2023)]{openpyxl}
Eric Gazoni and contributors.
\newblock openpyxl: A python library to read/write excel 2010 xlsx/xlsm files, 2023.

\bibitem[Goldblum et~al.(2024)Goldblum, Souri, Ni, Shu, Prabhu, Somepalli, Chattopadhyay, Ibrahim, Bardes, Hoffman, et~al.]{goldblum2024battle}
Micah Goldblum, Hossein Souri, Renkun Ni, Manli Shu, Viraj Prabhu, Gowthami Somepalli, Prithvijit Chattopadhyay, Mark Ibrahim, Adrien Bardes, Judy Hoffman, et~al.
\newblock Battle of the backbones: A large-scale comparison of pretrained models across computer vision tasks.
\newblock \emph{Advances in Neural Information Processing Systems}, 36, 2024.

\bibitem[Hipp(2020)]{sqlite2020hipp}
Richard~D Hipp.
\newblock {SQLite}, 2020.

\bibitem[Hunter and the Matplotlib Development~Team(2007)]{matplotlib}
John~D. Hunter and the Matplotlib Development~Team.
\newblock Matplotlib: A 2d graphics environment, 2007.

\bibitem[Hutter et~al.(2019)Hutter, Kotthoff, and Vanschoren]{Hutter2019}
Frank Hutter, Lars Kotthoff, and Joaquin Vanschoren.
\newblock \emph{Automated Machine Learning: Methods, Systems, Challenges}.
\newblock Springer, 2019.

\bibitem[Krizhevsky(2009)]{cifar}
Alex Krizhevsky.
\newblock Learning multiple layers of features from tiny images.
\newblock 2009.

\bibitem[Krizhevsky and Hinton(2009)]{krizhevsky2009learning}
Alex Krizhevsky and Geoffrey Hinton.
\newblock Learning multiple layers of features from tiny images.
\newblock Technical Report~0, University of Toronto, Toronto, Ontario, 2009.

\bibitem[Krizhevsky et~al.(2012)Krizhevsky, Sutskever, and Hinton]{NIPS2012_c399862d}
Alex Krizhevsky, Ilya Sutskever, and Geoffrey~E Hinton.
\newblock Imagenet classification with deep convolutional neural networks.
\newblock In \emph{Advances in Neural Information Processing Systems}. Curran Associates, Inc., 2012.

\bibitem[Lin et~al.(2015)Lin, Maire, Belongie, Bourdev, Girshick, Hays, Perona, Ramanan, Zitnick, and Dollár]{lin2015microsoftcococommonobjects}
Tsung-Yi Lin, Michael Maire, Serge Belongie, Lubomir Bourdev, Ross Girshick, James Hays, Pietro Perona, Deva Ramanan, C.~Lawrence Zitnick, and Piotr Dollár.
\newblock Microsoft coco: Common objects in context, 2015.

\bibitem[Lin et~al.(2017)Lin, Goyal, Girshick, He, and Dollár]{Lin2018}
Tsung-Yi Lin, Priya Goyal, Ross Girshick, Kaiming He, and Piotr Dollár.
\newblock Focal loss for dense object detection.
\newblock In \emph{Proceedings of the IEEE International Conference on Computer Vision (ICCV)}, pages 2980--2988, 2017.

\bibitem[Maas et~al.(2011)Maas, Daly, Pham, Huang, Ng, and Potts]{maas2011learning}
Andrew Maas, Raymond~E Daly, Peter~T Pham, Dan Huang, Andrew~Y Ng, and Christopher Potts.
\newblock Learning word vectors for sentiment analysis.
\newblock In \emph{Proceedings of the 49th annual meeting of the association for computational linguistics: Human language technologies}, pages 142--150, 2011.

\bibitem[McGarvey et~al.(2019)McGarvey, Nightingale, Luo, Huang, Martin, Wu, and Consortium]{uniprot}
Peter~B McGarvey, Andrew Nightingale, Jie Luo, Hongzhan Huang, Maria~J Martin, Cathy Wu, and UniProt Consortium.
\newblock Uniprot genomic mapping for deciphering functional effects of missense variants.
\newblock \emph{Human mutation}, 40\penalty0 (6):\penalty0 694--705, 2019.

\bibitem[McLaughlin et~al.(2015)]{McLaughlin2015}
Shane McLaughlin et~al.
\newblock A systematic review of software maintainability measurement.
\newblock \emph{Software: Practice and Experience}, 45\penalty0 (9):\penalty0 1281--1308, 2015.

\bibitem[Models(2022)]{dl-models}
Deep~Learning Models.
\newblock Deep learning models, 2022.

\bibitem[{ONNX Community}()]{ONNXModelZoo}
{ONNX Community}.
\newblock Onnx model zoo.
\newblock \url{https://github.com/onnx/models}.
\newblock Accessed: 2025-02-25.

\bibitem[pandas~development team(2020)]{pandas}
The pandas~development team.
\newblock pandas-dev/pandas: Pandas, 2020.

\bibitem[Paszke et~al.(2019)Paszke, Gross, Massa, Lerer, Bradbury, Chanan, Killeen, Lin, Gimelshein, Antiga, Desmaison, Köpf, Yang, DeVito, Raison, Tejani, Chilamkurthy, Steiner, Fang, Bai, and Chintala]{PyTorch}
Adam Paszke, Sam Gross, Francisco Massa, Adam Lerer, James Bradbury, Gregory Chanan, Trevor Killeen, Zeming Lin, Natalia Gimelshein, Luca Antiga, Alban Desmaison, Andreas Köpf, Edward Yang, Zach DeVito, Martin Raison, Alykhan Tejani, Sasank Chilamkurthy, Benoit Steiner, Lu Fang, Junjie Bai, and Soumith Chintala.
\newblock Pytorch: An imperative style, high-performance deep learning library, 2019.

\bibitem[Repository(2018)]{wolfram-models}
Wolfram Neural~Net Repository.
\newblock Wolfram neural net repository, 2018.

\bibitem[Rupani et~al.(2025)Rupani, Ignatov, and Timofte]{Rupani2025llm}
Bhavya Rupani, Dmitry Ignatov, and Radu Timofte.
\newblock Exploring the collaboration between vision models and llms for enhanced image classification.
\newblock \emph{Dimensions}, 27\penalty0 (1), 2025.
\newblock doi:10.13140/RG.2.2.14615.69284.

\bibitem[Russakovsky et~al.(2015)Russakovsky, Deng, Su, Krause, Satheesh, Ma, Huang, Karpathy, Khosla, Bernstein, Berg, and Fei-Fei]{ImageNet}
Olga Russakovsky, Jia Deng, Hao Su, Jonathan Krause, Sanjeev Satheesh, Sean Ma, Zhiheng Huang, Andrej Karpathy, Aditya Khosla, Michael Bernstein, Alexander~C. Berg, and Li Fei-Fei.
\newblock Imagenet large scale visual recognition challenge, 2015.

\bibitem[Schürholt et~al.(2022)Schürholt, Taskiran, Knyazev, i~Nieto, and Borth]{modelzoos}
Konstantin Schürholt, Diyar Taskiran, Boris Knyazev, Xavier~Giró i Nieto, and Damian Borth.
\newblock Model zoos: A dataset of diverse populations of neural network models, 2022.

\bibitem[Sweeney et~al.(2018)Sweeney, Azad, Donato, Haynes, Perumal, Henao, Bermejo-Martin, Almansa, Tamayo, Howrylak, et~al.]{sweeney2018unsupervised}
Timothy~E Sweeney, Tej~D Azad, Michele Donato, Winston~A Haynes, Thanneer~M Perumal, Ricardo Henao, Jes{\'u}s~F Bermejo-Martin, Raquel Almansa, Eduardo Tamayo, Judith~A Howrylak, et~al.
\newblock Unsupervised analysis of transcriptomics in bacterial sepsis across multiple datasets reveals three robust clusters.
\newblock \emph{Critical care medicine}, 46\penalty0 (6):\penalty0 915--925, 2018.

\bibitem[Wang et~al.(2019)Wang, Singh, Michael, Hill, Levy, and Bowman]{GLUE}
Alex Wang, Amanpreet Singh, Julian Michael, Felix Hill, Omer Levy, and Samuel~R. Bowman.
\newblock Glue: A multi-task benchmark and analysis platform for natural language understanding, 2019.

\bibitem[Waskom and the Seaborn Development~Team(2023)]{seaborn}
Michael Waskom and the Seaborn Development~Team.
\newblock Seaborn: Statistical data visualization, 2023.

\bibitem[Wolf et~al.(2020)Wolf, Debut, Sanh, Chaumond, Delangue, Moi, Cistac, Rault, Louf, Funtowicz, Davison, Shleifer, von Platen, Ma, Jernite, Plu, Xu, Scao, Gugger, Drame, Lhoest, and Rush]{huggingface}
Thomas Wolf, Lysandre Debut, Victor Sanh, Julien Chaumond, Clement Delangue, Anthony Moi, Pierric Cistac, Tim Rault, Rémi Louf, Morgan Funtowicz, Joe Davison, Sam Shleifer, Patrick von Platen, Clara Ma, Yacine Jernite, Julien Plu, Canwen Xu, Teven~Le Scao, Sylvain Gugger, Mariama Drame, Quentin Lhoest, and Alexander~M. Rush.
\newblock Huggingface's transformers: State-of-the-art natural language processing, 2020.

\bibitem[Yang et~al.(2024)Yang, Gao, Peng, Huang, Tang, and Zhan]{younger}
Zhengxin Yang, Wanling Gao, Luzhou Peng, Yunyou Huang, Fei Tang, and Jianfeng Zhan.
\newblock Younger: The first dataset for artificial intelligence-generated neural network architecture, 2024.

\bibitem[Zoph et~al.(2018)Zoph, Vasudevan, Shlens, and Le]{Zoph2018}
Barret Zoph, Vijay Vasudevan, Jonathon Shlens, and Quoc~V. Le.
\newblock Learning transferable architectures for scalable image recognition.
\newblock In \emph{Proceedings of the IEEE Conference on Computer Vision and Pattern Recognition (CVPR)}, pages 8697--8710, 2018.

\bibitem[Zvyagin et~al.(2021)Zvyagin, Brettin, Ramanathan, and Jha]{crossedwires}
Max Zvyagin, Thomas Brettin, Arvind Ramanathan, and Sumit~Kumar Jha.
\newblock Crossedwires: A dataset of syntactically equivalent but semantically disparate deep learning models, 2021.

\end{thebibliography}
}

\clearpage
\setcounter{page}{1}
\maketitlesupplementary

\section{SQL Database}
\label{SQL}
We chose SQLite \cite{sqlite2020hipp} for its lightweight, portable, and serverless design, ensuring seamless distribution and easy integration. Its schema-based structure ensures data integrity, making it ideal for managing tasks, datasets, metrics, and results. SQLite excels at relational queries like filtering and trend analysis while offering efficient joins, aggregations, and data handling, which support scalability for analyzing neural network performance. Its simplicity enables effortless setup with Python libraries like Pandas, and its zero-configuration design minimizes administrative overhead, allowing users to focus on experiments. These features make SQLite an optimal choice for this framework.

Our provided SQL database comprises multiple tables designed to capture distinct aspects of the system's functionality and ensure efficient data management. These tables are categorized into three primary groups: Main Table, Code Tables, and Parameter Table.
\\
\textbf{Main Table:} The stat table is dedicated to storing performance statistics, including key metrics such as accuracy and duration. These records provide a detailed overview of model behavior during both the training and testing phases, offering critical insights into the effectiveness of various configurations. 
\\
\textbf{Code Tables: } The code tables comprise three key entities: nn, which documents the neural network configurations, including architecture and layer details; metric, which retains information about evaluation metrics such as accuracy or intersection over union (IoU); and transform, which records data transformation techniques applied during preprocessing.
\\
\textbf{Parameter Table:} To ensure reproducibility, parameter configurations are maintained in the prm table, which includes hyperparameters such as learning rates, batch sizes, and momentum values. This comprehensive structure facilitates the seamless reproduction and comparison of experimental setups.

The design of the database schema is further enhanced by a carefully considered column structure. Index columns, including key identifiers such as task, dataset, and dependent columns (e.g., metric and transform), are implemented to enable efficient indexing and filtering. Additionally, metrics columns capture critical performance indicators such as accuracy, duration, and other essential results, forming the basis for monitoring and evaluating model outcomes.

The relationships between tables are established through foreign key constraints, ensuring data integrity and enabling the execution of complex queries that span tasks, metrics, and results. This interconnected structure fosters traceability, allowing data configurations to be directly linked to their corresponding results, thereby supporting comprehensive analysis.

The database schema is designed with scalability and flexibility in mind, accommodating future extensions, such as new tasks, datasets, or performance metrics, without necessitating significant redesigns. Dynamic indexing mechanisms are incorporated to maintain query efficiency as the database size expands. This robust and adaptable design ensures that the database remains a reliable and efficient component of the framework, capable of supporting a wide range of experimental and analytical needs.

\section{Details of The Supported Deep Learning Tasks}
\label{dl-task-details}
This section provides a detailed breakdown of the image classification, image segmentation, and object detection tasks supported by the NN Dataset framework. For each task, we describe the dataset loaders, preprocessing transformations, evaluation metrics, and hyperparameter configurations that enable standardized training and evaluation.

We outline the datasets used, the still evolving \cite{Aboudeshish2025augmentation} preprocessing techniques such as normalization and augmentation, and the evaluation metrics—accuracy for classification, mean IoU for segmentation, and mAP for object detection. Additionally, we detail the hyperparameter configurations, including learning rates, batch sizes, and task-specific settings like anchor box sizes. The integration of Optuna for automated hyperparameter tuning is also covered, demonstrating how the framework optimizes model performance efficiently.

The following subsections provide a structured, task-by-task explanation of these components.
\subsection{Image Classification}
\label{classification}
One of the main deep learning tasks that is supported by the neural network dataset is image classification.
Several data loaders and transformations for the most well known datasets have been included. 
The most prominent datasets included for image classification are MNIST \cite{mnist}, CIFAR-10 and CIFAR-100 \cite{cifar}.

Additionally, the framework allows easy switching between transformations such as \texttt{NormalizeToFloat}, \texttt{ToComplex64}, or other custom preprocessing steps. Different types of image transformations are used to evaluate model performance. For instance, the \texttt{NormalizeToFloat} transformation converts input data into a \texttt{float32} NumPy array and scales pixel values to the range [0, 1], stabilizing the training process by ensuring consistent input ranges. Standard preprocessing steps, including resizing, cropping, flipping, and normalizing images, are also integrated into the pipeline, ensuring datasets are uniformly prepared for neural network training. This automated and flexible design minimizes manual effort while supporting diverse datasets and applications, making the framework both accessible and efficient.

The metric provided to evaluate the performance of image classification models is accuracy, which is the most intuitive metric for classification tasks. This metric calculates the number of true predictions and compares them to the total number of tested data points.

An example of the command to run a task for image classification is written in \cref{listing:configuration_code}.

\subsection{Image Segmentation}
\label{segmentation}

Another DL task suppoted by the NN-Dataset framework is image segmentation. Our datasets includes a various number of well-known neural network models for this task. 
Many of these models rely on different classification backbones and are compatible with several different ones. 
To support these varieties, we have provided different implementations of these models with hard coded back bones.

The included dataset for image classification is COCO-Seg 2017 \cite{lin2015microsoftcococommonobjects}. A COCO dataset loader processes annotations to generate masks, with categories reduced to 6 by default to improve training accuracy and speed, as most segmentation models use an image classification backbone for feature extraction \cite{goldblum2024battle}. The loader supports all 91 categories for state-of-the-art comparisons but generating masks remains a bottleneck due to computational demands. To improve efficiency, we use two worker threads for sampling, as pre-storing masks is impractical.

The dataset loader also supports category limitation and a preprocessing step to filter samples with insufficient mask pixels. Since this filtering slows evaluation, we store accepted image IDs and filtering parameters in a temporary file to skip redundant processing. Additionally, the loader limits the maximum sample size, though small limits degrade performance. Resizing is handled separately: bilinear interpolation for images and nearest-neighbor interpolation for masks, avoiding artifacts and skipping normalization for masks.

For evaluation, the mean IoU metric is used instead of pixel accuracy. Mean IoU is integrated into the metrics provided with the framework, supporting batched calculations and using the default COCO validation dataset for consistent model comparisons.

As a side note, to emphasize how our framework can be used for performance analysis, we have noticed that simpler backbones outperformed more complex ones (e.g., FCN32s with ResNet50 achieved 0.439 mean IoU, while ResNet101 achieved only 0.400 after five epochs). Pre-trained backbones are expected to improve results but were excluded as they are readily available online. Sensitivity to hyperparameter tuning and difficulty in training feature extraction layers limited model performance.

\subsection{Object Detection}
Object detection is a downstream task in the pipeline, aiming to place bounding boxes around objects in images. These boxes are typically labeled using X and Y coordinates or a single corner with height and width. Unlike classification, this task involves both object identification and localization.

Following the segmentation dataset approach, the COCO 2017 dataset is automatically downloaded and extracted if unavailable. Category reduction is applied, and images without relevant objects are discarded.

Batch preparation requires special handling due to multiple bounding boxes per image. Unlike classification, where images and labels can be stacked, detection labels vary in number and cannot be stacked directly. To ensure compatibility with the framework, labels are stored as dictionaries and processed using a custom collate function, inspired by torchvision. This method maintains interoperability while avoiding inefficiencies like fixed-format batching.

Transformations such as resizing, cropping, and rotating must preserve label accuracy. To simplify this, images are padded to a fixed size, ensuring bounding boxes remain unchanged and compatible with the pipeline.

For training, detection losses are handled as a variable-length dictionary to accommodate classification, localization, and auxiliary losses. This allows flexible weighting for precise localization or identification using the learn and trainsetup functions for each neural network individually.

Pretrained backbone networks are used to reduce computational overhead and focus optimization on the detection network, aligning with the pipeline’s time constraints.

The primary evaluation metric is mean Average Precision (mAP). While COCO averages over multiple IoU thresholds, a default threshold of 0.5 is used here for quicker hyperparameter tuning before further refinement.

\section{Determined Accuracies}
In this section, we provide the determined accuracies for the included models in the LEMUR dataset. 
These are the best accuracies achieved by each model during hyperparameter tuning with optuna.

\begin{table}[t]
\small
\centering
\caption{Comparison of classification models with the total number of their parameters (Params.) in millions, input image resolution (Res.) in pixels, and accuracy metric yielding the best results by Optuna on CIFAR-10 dataset within the LEMUR framework}
\label{tab:classification}
\begin{tabular*}{\columnwidth}{@{\extracolsep{\fill}}lccc@{}}
\toprule
\textbf{Model} & \textbf{Params. ($\times10^6$)} & \textbf{Res.} (px) & \textbf{Accuracy} \\
\midrule
AirNet & 4.91 & 32$\times$32 & 0.8077 \\
AirNext & 1.51 & 256$\times$256 & 0.7769 \\
AlexNet & 57.04 & 299$\times$299 & 0.8675 \\
BagNet & 1.25 & 512$\times$512 & 0.6824 \\
BayesianNet-1 & 3.57 & 32$\times$32 & 0.6710 \\
BayesianNet-2 & 4.35 & 32$\times$32 & 0.6046 \\
BayesianNet-3 & 0.12 & 32$\times$32 & 0.5509 \\
ComplexNet & 0.52 & 32$\times$32 & 0.7268 \\
ConvNeXt & 49.46 & 32$\times$32 & 0.5617 \\
DPN107 & 0.02 & 32$\times$32 & 0.7340 \\
DPN131 & 0.06 & 32$\times$32 & 0.7685 \\
DPN68 & 0.06 & 32$\times$32 & 0.6530 \\
DarkNet & 3.66 & 32$\times$32 & 0.8499 \\
DenseNet & 25.53 & 128$\times$128 & 0.8792 \\
EfficientNet & 4.02 & 256$\times$256 & 0.9274 \\
GoogLeNet & 9.96 & 299$\times$299 & 0.9182 \\
ICNet & 0.05 & 256$\times$256 & 0.7166 \\
InceptionV3-1 & 21.81 & 512$\times$512 & 0.8606 \\
InceptionV3-2 & 24.37 & 512$\times$512 & 0.8665 \\
MNASNet & 1.9 & 299$\times$299 & 0.8280 \\
MaxVit & 30.38 & 299$\times$299 & 0.8813 \\
MobileNetV2 & 2.24 & 299$\times$299 & 0.8661 \\
MobileNetV3 & 0.59 & 512$\times$512 & 0.8688 \\
RegNet & 3.91 & 299$\times$299 & 0.8495 \\
ResNet & 11.18 & 299$\times$299 & 0.8370 \\
ShuffleNet & 1.26 & 299$\times$299 & 0.8441 \\
SqueezeNet-1 & 0.74 & 299$\times$299 & 0.8005 \\
SqueezeNet-2 & 0.73 & 32$\times$32 & 0.6913 \\
SwinTransformer & 27.53 & 32$\times$32 & 0.7482 \\
UNet2D & 0.56 & 32$\times$32 & 0.5796 \\
VGG & 128.81 & 299$\times$299 & 0.8242 \\
VisionTransformer & 85.23 & 32$\times$32 & 0.4885 \\
\bottomrule
\end{tabular*}
\end{table}

\begin{table}[t]
\small
\centering
\caption{
Comparison of segmentation models with the total number of their parameters (Params.) in millions, input image resolution (Res.) in pixels, and mean intersection over union (mIoU) metric yielding the best results by Optuna on CIFAR-10 dataset within the LEMUR framework
}
\label{tab:segmentation}
\begin{tabular*}{\columnwidth}{@{\extracolsep{\fill}}lccc@{}}
\toprule
\textbf{Model} & \textbf{Params. ($\times10^6$)} & \textbf{Res.} (px) & \textbf{mIoU} \\
\midrule
DeepLabV3-1 & 39.84 & 64$\times$64 & 0.3731 \\
DeepLabV3-2 & 58.83 & 128$\times$128 & 0.3543 \\
FCN16s & 15.31 & 256$\times$256 & 0.4365 \\
FCN32s-1 & 32.95 & 512$\times$512 & 0.4386 \\
FCN32s-2 & 51.94 & 128$\times$128 & 0.3200 \\
FCN32s-3 & 15.31 & 512$\times$512 & 0.4101 \\
FCN8s & 15.31 & 64$\times$64 & 0.3020 \\
LRASPP & 3.22 & 256$\times$256 & 0.3810 \\
UNet-1 & 31.04 & 128$\times$128 & 0.4494 \\
UNet-2 & 17.26 & 64$\times$64 & 0.3393 \\
\bottomrule
\end{tabular*}
\end{table}

\begin{table}[t]
\small
\centering
\caption{Comparison of classification models with the total number of their parameters (Params.) in millions, input image resolution (Res.) in pixels, and mean average precision metric at 0.5 intersection over union (mAP), yielding the best results by Optuna on CIFAR-10 dataset within the LEMUR framework}
\label{tab:detection}
\begin{tabular*}{\columnwidth}{@{\extracolsep{\fill}}lccc@{}}
\toprule
\textbf{Model} & \textbf{Params. ($\times10^6$)} & \textbf{Res.} (px) & \textbf{mAP} \\
\midrule
FCOS & 32.13 & 800$\times$1333 & 0.7394 \\
FasterRCNN & 41.37 & 800$\times$1333 & 0.6214 \\
RetinaNet & 32.31 & 800$\times$1333 & 0.0922 \\
SSDLite & 3.17 & 300$\times$300 & 0.3975 \\
\bottomrule
\end{tabular*}

\end{table}

\cref{tab:classification} highlights the results of the included classification models while \cref{tab:segmentation} and \cref{tab:detection} show the results of segmentation and detection models respectively.

\section{Experimental Setup for Optuna}
This section details the hyperparameter optimization process using Optuna, including the search space definitions, tuning parameters, and evaluation criteria. We outline the selected hyperparameter ranges and the strategies used for efficient model optimization across image classification, segmentation, and object detection tasks.

Hyperparameters are dynamically specified prior to initiating training, enhancing the flexibility and adaptability of the training process. Users can define hyperparameter values or ranges directly through command-line parameters, enabling seamless integration with diverse neural network configurations and simplifying reproducibility.

The key hyperparameters include:
\begin{itemize}
    \item \textbf{Learning Rate}: Defined as a logarithmic range (e.g., \texttt{--min\_learning\_rate} and \texttt{--max\_learning\_rate}) to allow fine-grained control over optimization dynamics.
    \item \textbf{Batch Size}: Specified as a binary power range (e.g., \texttt{--min\_batch\_binary\_power} and \texttt{--max\_batch\_binary\_power}) to streamline memory utilization and scalability across architectures.
    \item \textbf{Momentum}: Defined as a uniform range (e.g., \texttt{--min\_momentum} and \texttt{--max\_momentum}) to balance stability and convergence speed.
    \item \textbf{Transformations}: Configurable transformations (e.g., normalization, resizing) are passed as parameters (\texttt{--transform}), enabling targeted preprocessing for image-based tasks.
\end{itemize}

To enforce specific hyperparameters, minimum and maximum values for each hyperparameter can be set to identical values. 

For instance, training AlexNet \cite{NIPS2012_c399862d} with precise hyperparameters can be executed as shown in Listing~\ref{listing:configuration_code}.

Each model from the LEMUR framework was subjected to automated hyperparameter tuning using Optuna’s sampling methods. The process leveraged both fixed and dynamic configurations, ensuring consistency and adaptability. Training was conducted for every epoch from 1 to 50, with the results logged automatically at each step, providing a comprehensive dataset for detailed statistical analysis.

\section{Workflow of Visualization}
This section outlines the visualization workflow used to analyze model performance within the NN Dataset framework. We describe the data processing steps, statistical summaries, and plotting techniques used to generate insights across different tasks. Additionally, we detail how key metrics such as accuracy, IoU, and mAP are visualized through scatter plots, histograms, box plots, and correlation heatmaps to facilitate model comparison and analysis.

The statistical analysis workflow for the LEMUR framework is systematically organized into a sequence of clearly defined steps to ensure clarity, accuracy, and relevance. The structured workflow consists of the following stages:

\textbf{Data Acquisition:}
The process begins with the retrieval of raw data from the LEMUR framework through a well-defined API, utilizing the \texttt{fetch\_all\_data} function. This step forms the foundational stage of the workflow, providing the raw data required for subsequent processing and visualization. The acquired data acts as a critical link between the analytical workflow and the overarching objectives of the framework, ensuring alignment with the intended purpose of evaluating and optimizing neural network performance.

\textbf{Data Preparation:}
The data preparation process begins by applying the \texttt{filter\_raw\_data} module to remove irrelevant or noisy data from the dataset. This step ensures the retention of only essential columns, including \texttt{"task"}, \texttt{"dataset"}, \texttt{"metric"}, \texttt{"epoch"}, \texttt{"duration"}, \texttt{"accuracy"}, and \texttt{"nn"}. By narrowing the dataset to these critical variables, the framework establishes a foundation for precise and meaningful analysis.

Subsequently, the \texttt{process\_data} module is utilized, leveraging the robust functionality of the \texttt{pandas} library \cite{pandas}. The \texttt{pandas} library was chosen for its advanced data manipulation capabilities, enabling efficient grouping and the computation of key statistical measures. This module computes aggregated statistics, including the mean and standard deviation for critical metrics, providing a reliable assessment of model consistency. Such computations offer valuable insights into central tendencies and variability across training epochs.

These statistical measures play an essential role in identifying patterns and anomalies within the dataset. Patterns and trends provide opportunities to refine model performance, while anomalies may highlight areas requiring further investigation. By aligning closely with the objectives of the LEMUR framework, these insights contribute to enhancing the overall robustness and efficiency of the models under evaluation.

Moreover, task-specific analyses further refine the understanding of model behavior. This comprehensive approach, combining general statistical measures with domain-specific metrics, enables researchers to identify meaningful trends and actionable conclusions. By systematically addressing both global and task-specific concerns, this step provides a detailed and structured evaluation of neural network performance, fostering opportunities for continuous improvement.

\textbf{Visualization Generation: }
Leverage the Python libraries \texttt{matplotlib} \cite{matplotlib} and \texttt{seaborn} \cite{seaborn} for crafting sophisticated visualizations. These libraries were selected for their flexibility, user-friendly interfaces, and capability to generate publication-quality plots. The \texttt{matplotlib} library facilitates the customization of plot configurations and detailed chart formatting, while \texttt{seaborn} simplifies the creation of visually appealing statistical graphics, enhancing the clarity and interpretability of the results.

\section{Further Visualizations \& Insights}
In this section we provide further illustrations generated by the LEMUR framework on its set of neural networks and discuss the insights that we could draw from them.

To streamline the visualization process, the \texttt{generate\_plots.py} script is employed to produce a diverse range of visualizations tailored to specific research needs. In addition to the scatter plots discussed in \cref{visualization}, other visualization techniques incorporated into LEMUR include line and box plots. 

\begin{figure}[H]
    \centering
    \includegraphics[width=0.48\textwidth]{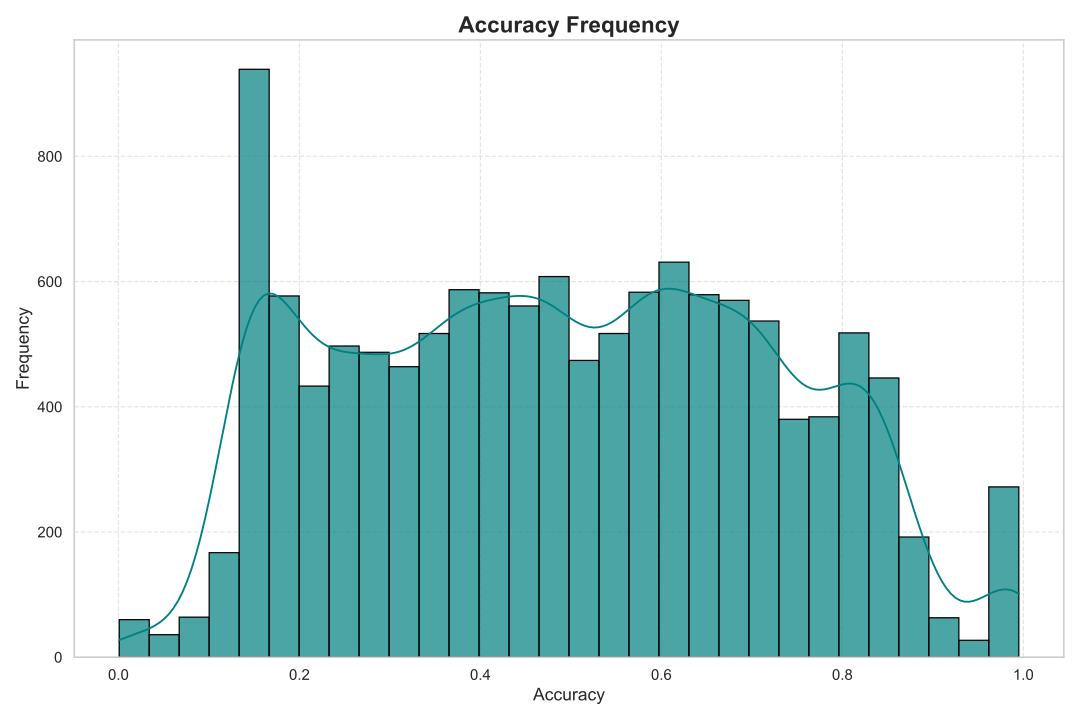}
    \caption{Histogram depicting the frequency distribution of accuracy values across all tasks. The plot reveals a bimodal distribution, with a significant concentration of values around 0.2 and 0.6, indicating common accuracy ranges achieved by models. The distribution suggests varying levels of model performance, with some achieving near-perfect accuracy while others cluster in lower accuracy ranges.}
    \label{fig:accuracy_histogram}
\end{figure}

\begin{figure}[H]
    \centering
    \includegraphics[width=0.48\textwidth]{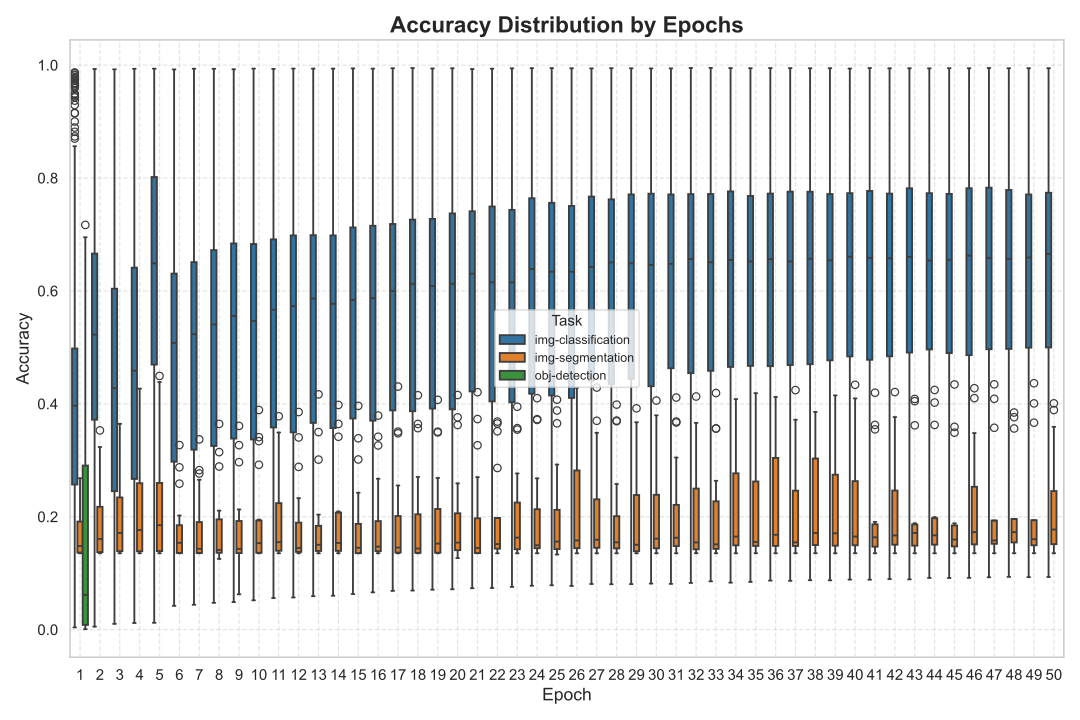}
    \caption{Box plot showing the distribution of accuracy across epochs for various tasks (image classification, image segmentation, and object detection). Image classification demonstrates a rapid increase in accuracy with minimal variability after the initial epochs, indicating stable convergence. Image segmentation and object detection show more gradual improvements and wider variability, reflecting the challenges of these tasks. The narrowing range over epochs highlights the models' stabilization as training progresses.}
    \label{fig:accuracy_vs_epochs_box}
\end{figure}

\begin{figure}[H]
    \centering
    \includegraphics[width=0.48\textwidth]{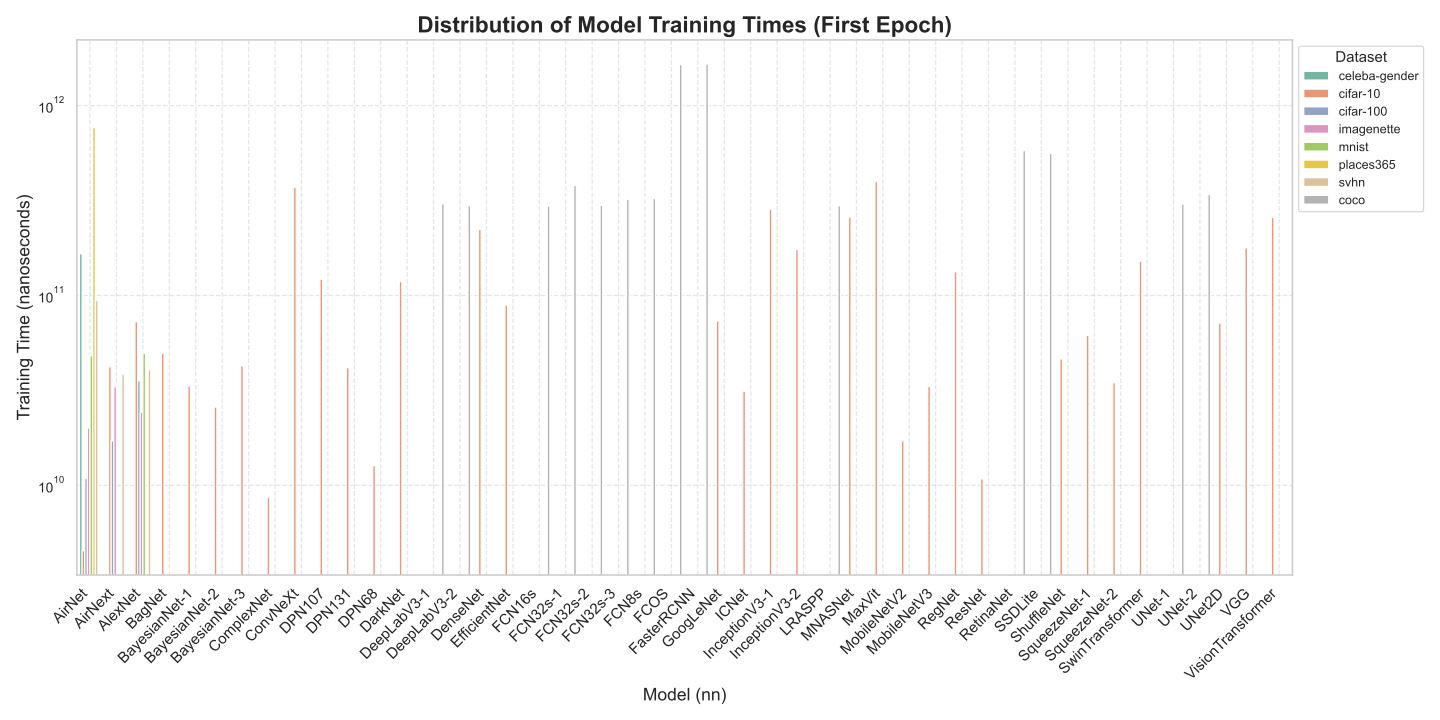}
    \caption{  Distribution of model training times (in nanoseconds) for the first epoch across various datasets and models. The plot illustrates the variability in computational requirements among different neural network architectures and datasets. Models trained on simpler datasets, such as MNIST, exhibit shorter training times, while complex datasets like COCO and Places365 require significantly longer times. This highlights the trade-offs between model complexity, dataset difficulty, and computational efficiency}
    \label{fig:first_epoch_training_time_distribution_by_model}
\end{figure}

\begin{figure*}[ht]
    \centering
    \begin{minipage}{0.33\linewidth}
        \centering
        \textit{(a)} \\
        \includegraphics[width=\linewidth]{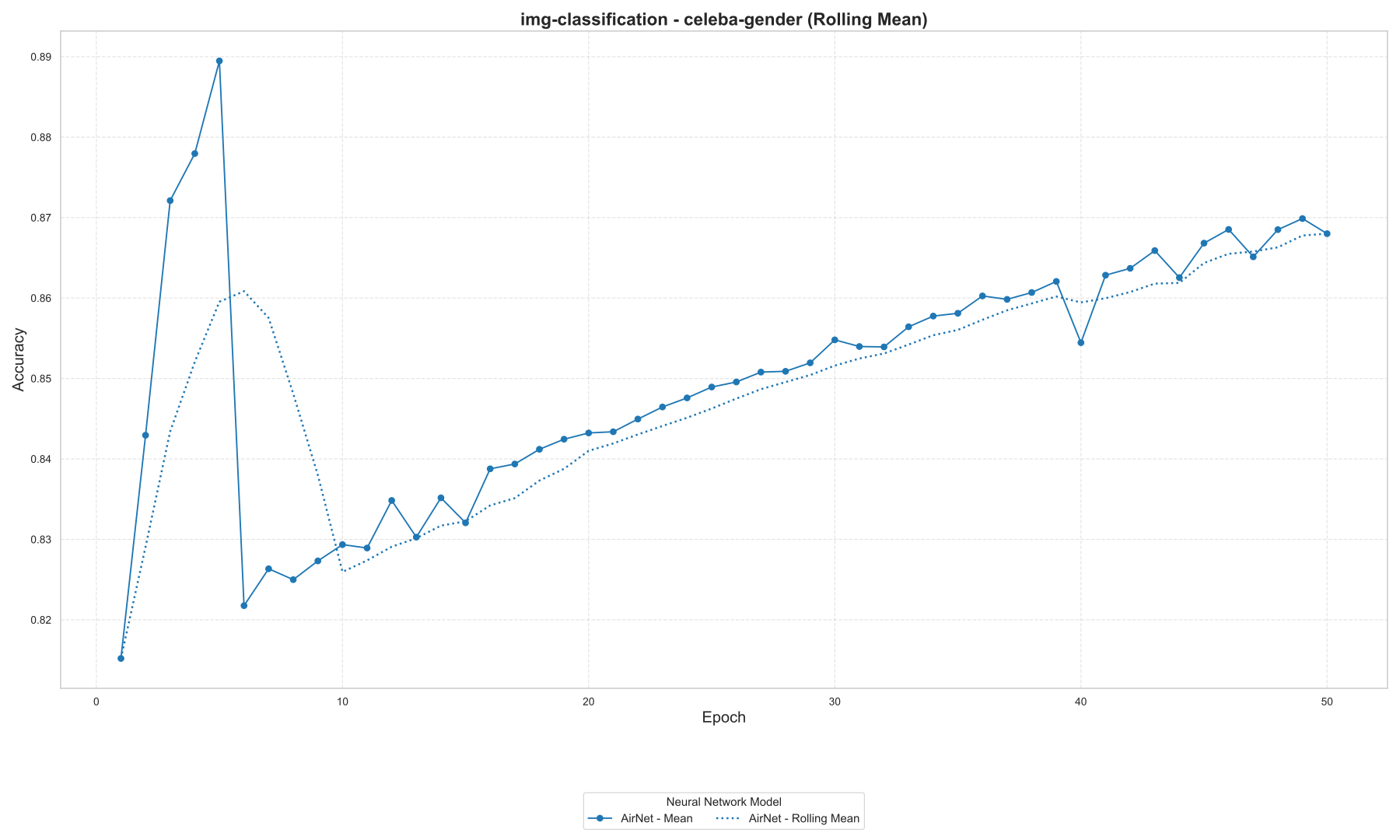}
    \end{minipage}
    \hfill
    \begin{minipage}{0.33\linewidth}
        \centering
        \textit{(b)} \\
        \includegraphics[width=\linewidth]{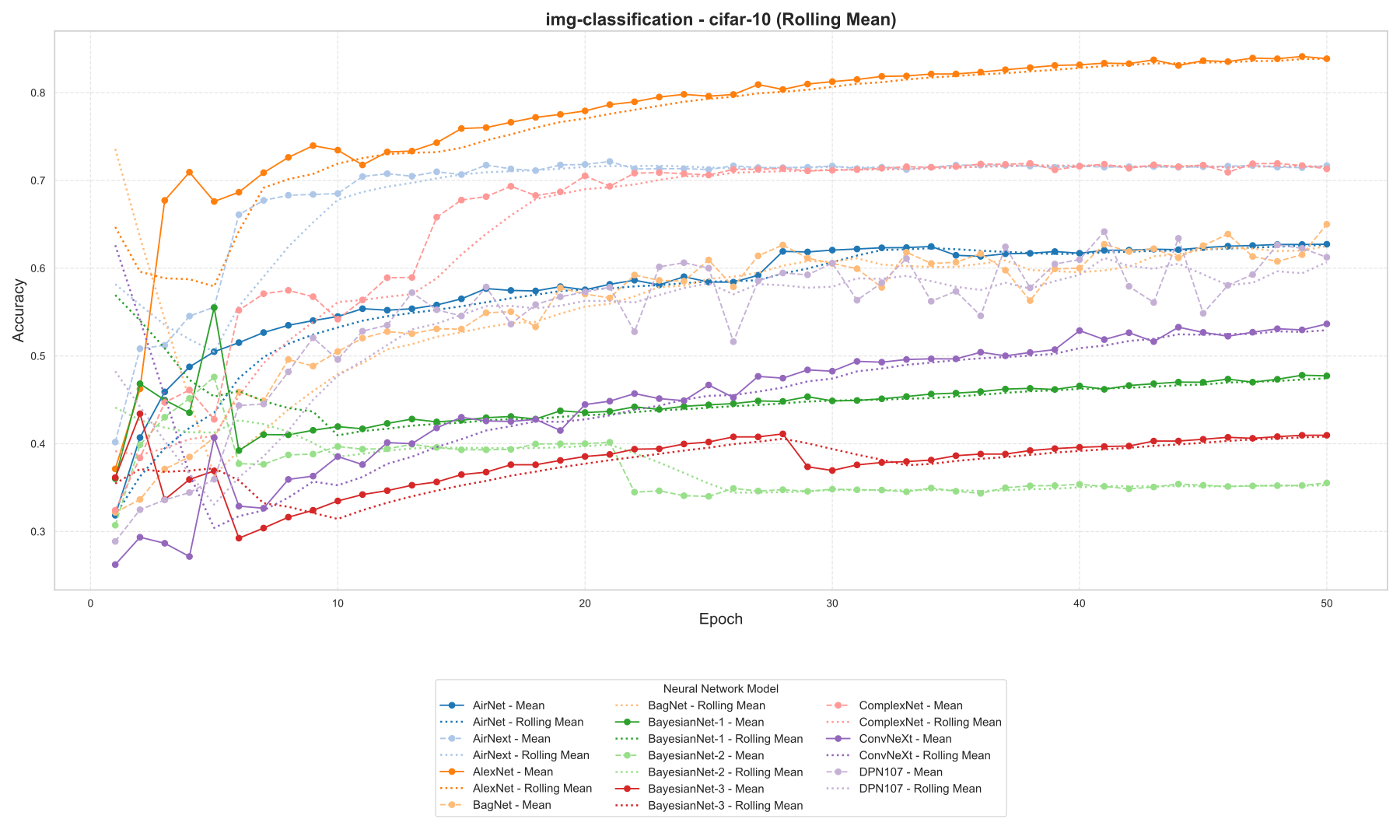}
    \end{minipage}
    \hfill
    \begin{minipage}{0.33\linewidth}
        \centering
        \textit{(c)} \\
        \includegraphics[width=\linewidth]{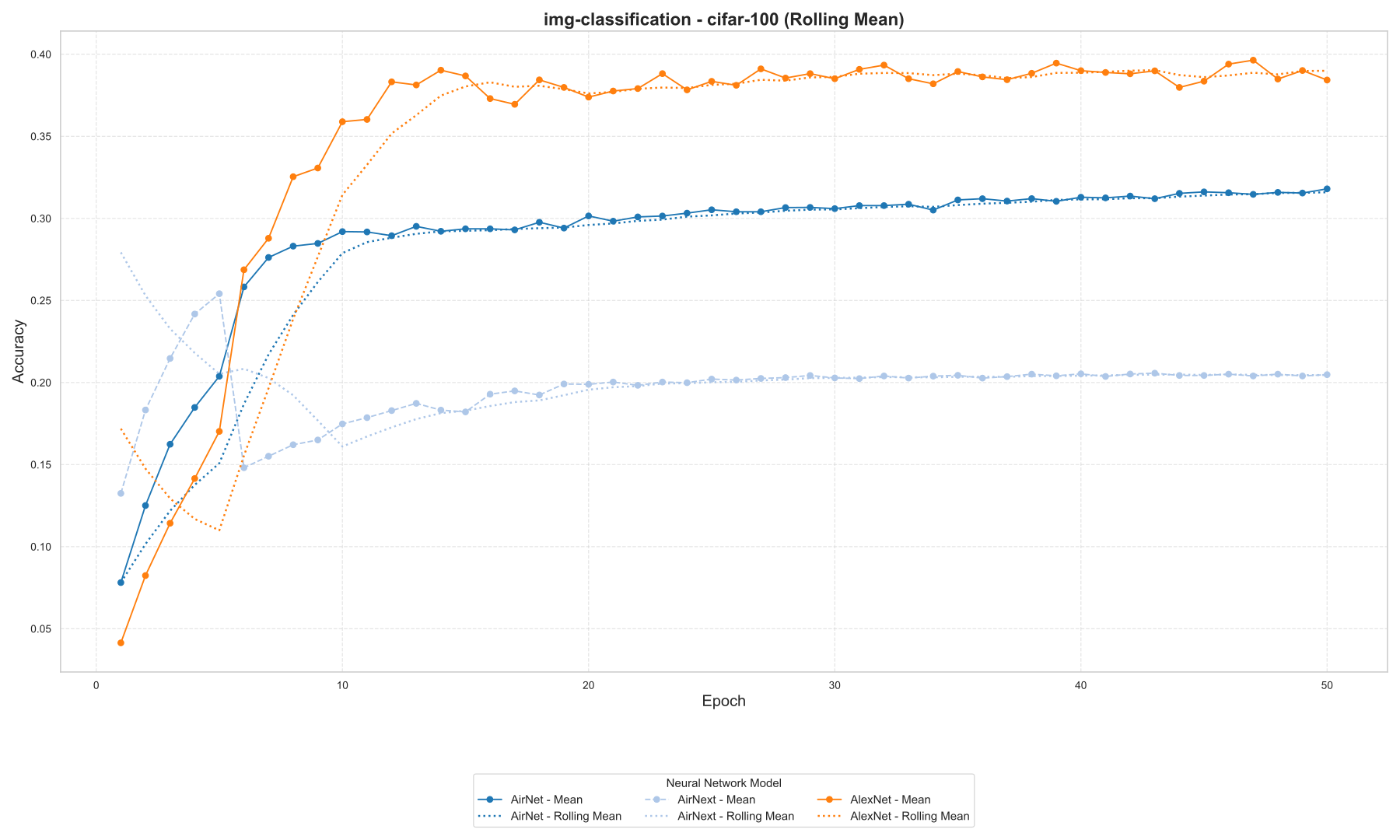}
    \end{minipage}
    \vskip 15pt
    \begin{minipage}{0.33\linewidth}
        \centering
        \textit{(d)} \\
        \includegraphics[width=\linewidth]{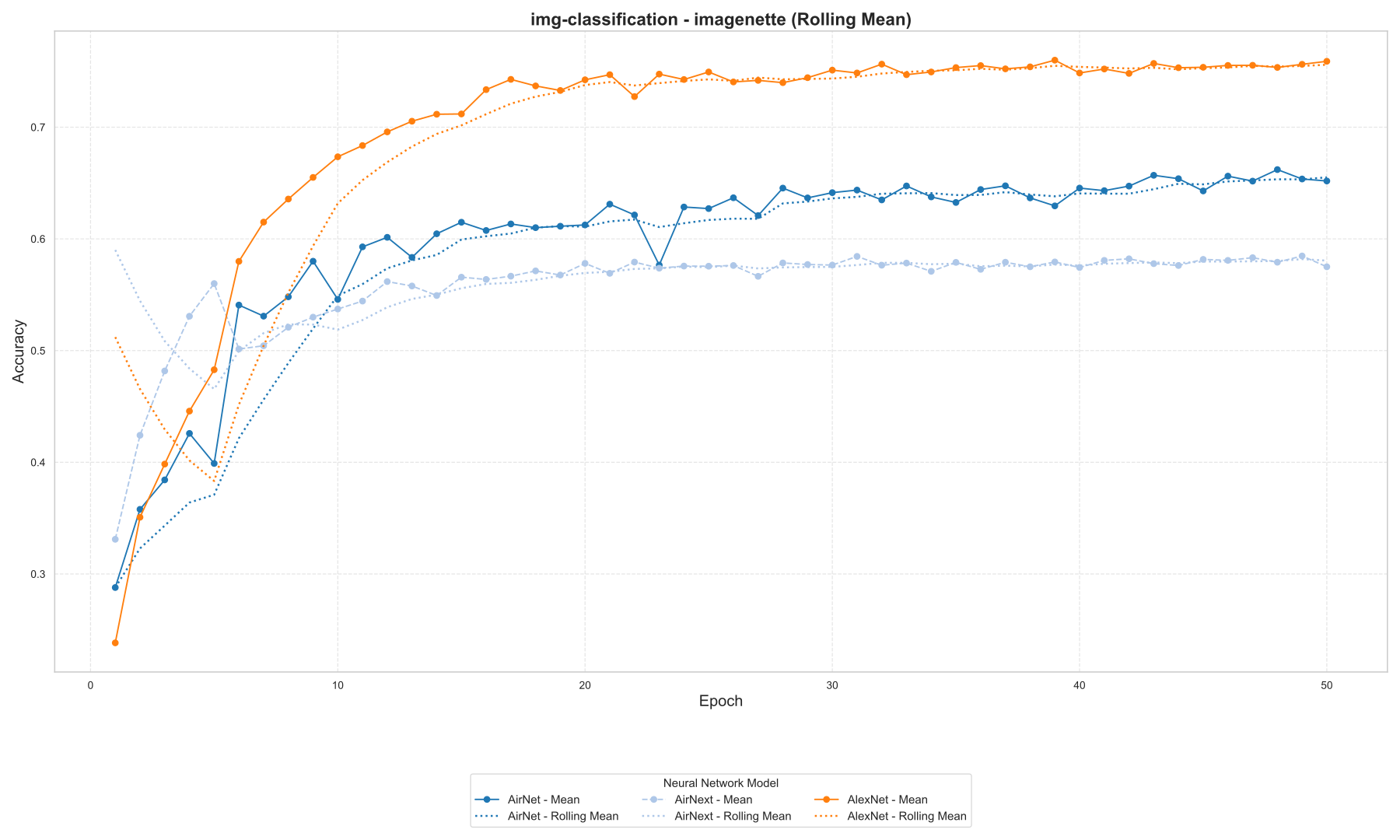}
    \end{minipage}
    \hfill
    \begin{minipage}{0.33\linewidth}
        \centering
        \textit{(e)} \\
        \includegraphics[width=\linewidth]{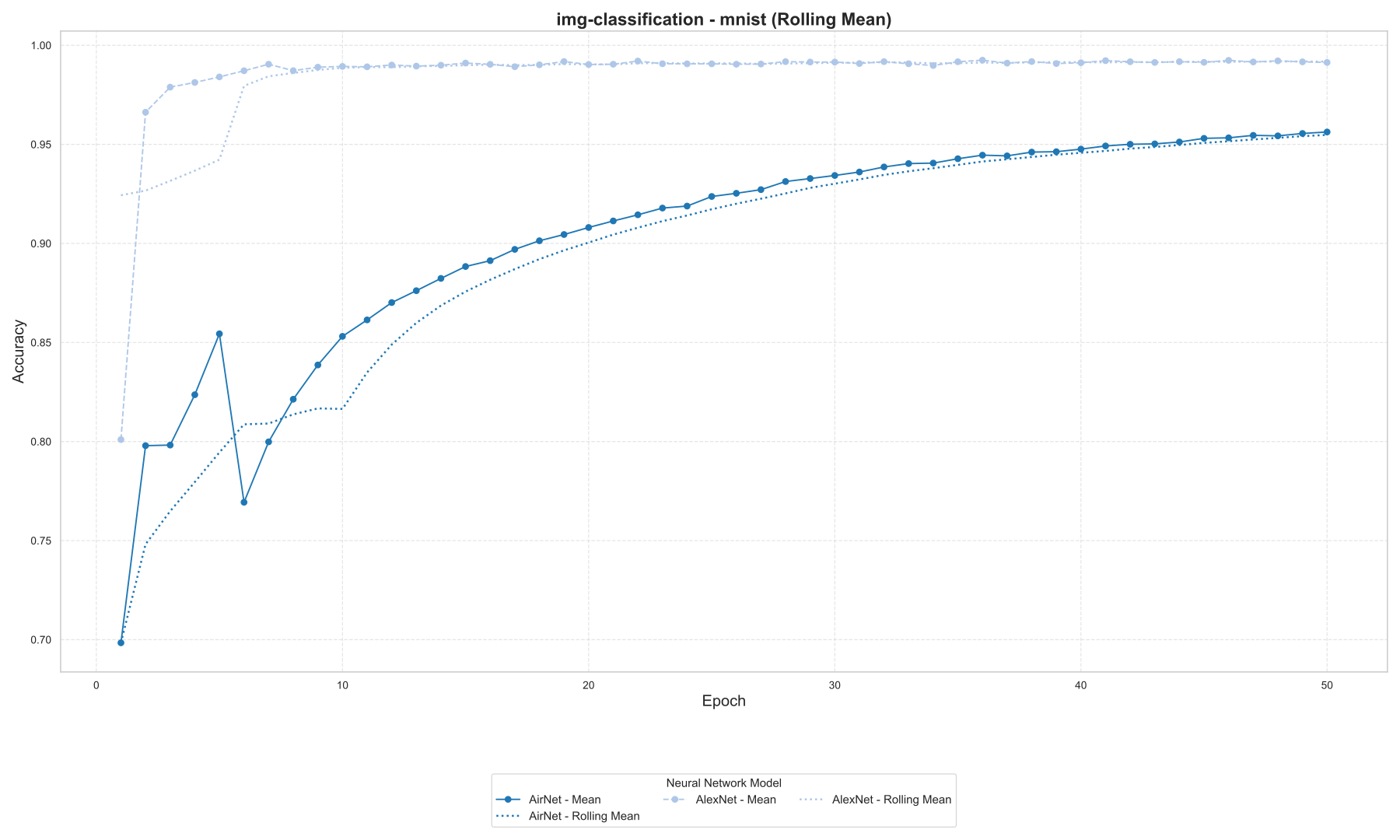}
    \end{minipage}
    \hfill
    \begin{minipage}{0.33\linewidth}
        \centering
        \textit{(f) } \\
        \includegraphics[width=\linewidth]{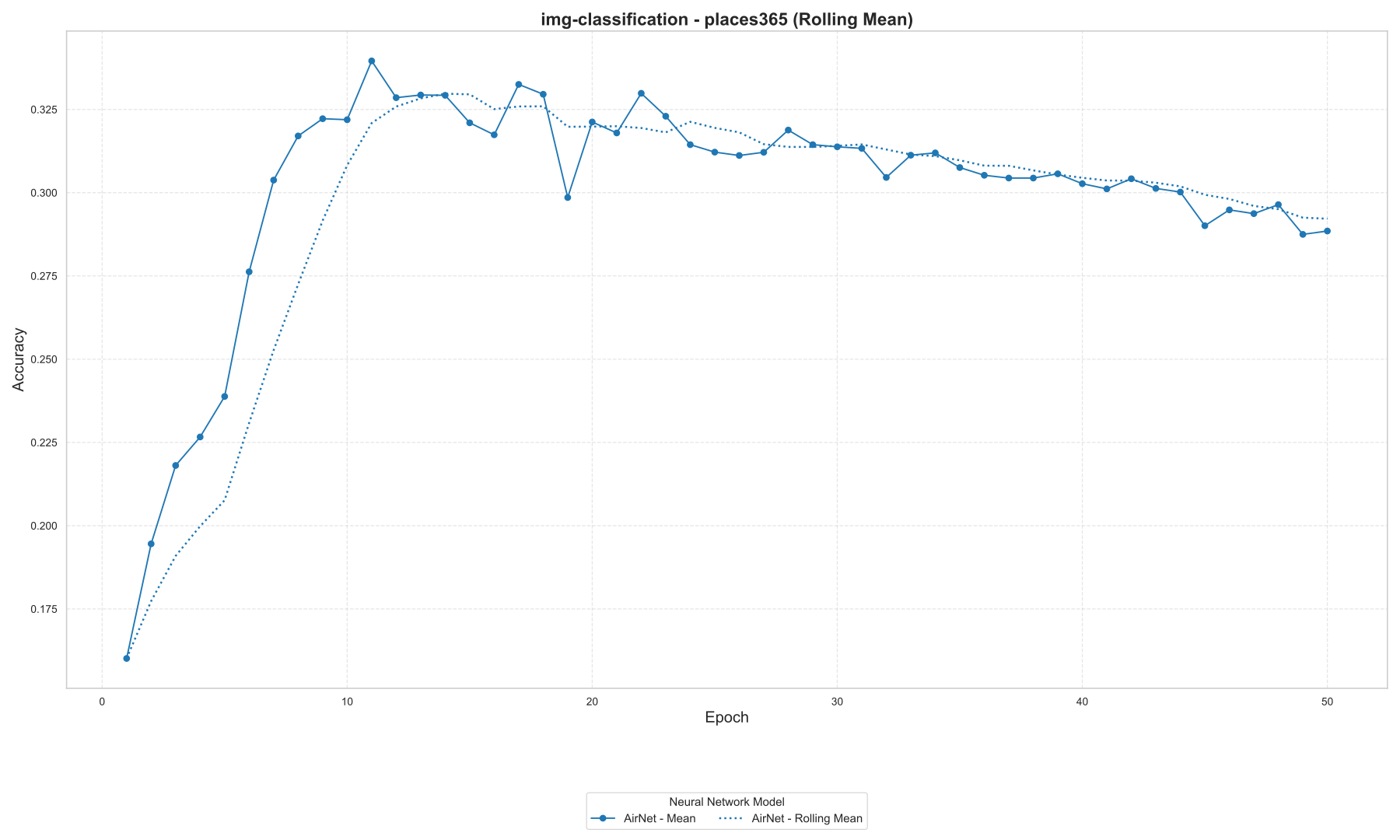}
    \end{minipage}
    \vskip 15pt
    \begin{minipage}{0.33\linewidth}
        \centering
        \textit{(g) } \\
        \includegraphics[width=\linewidth]{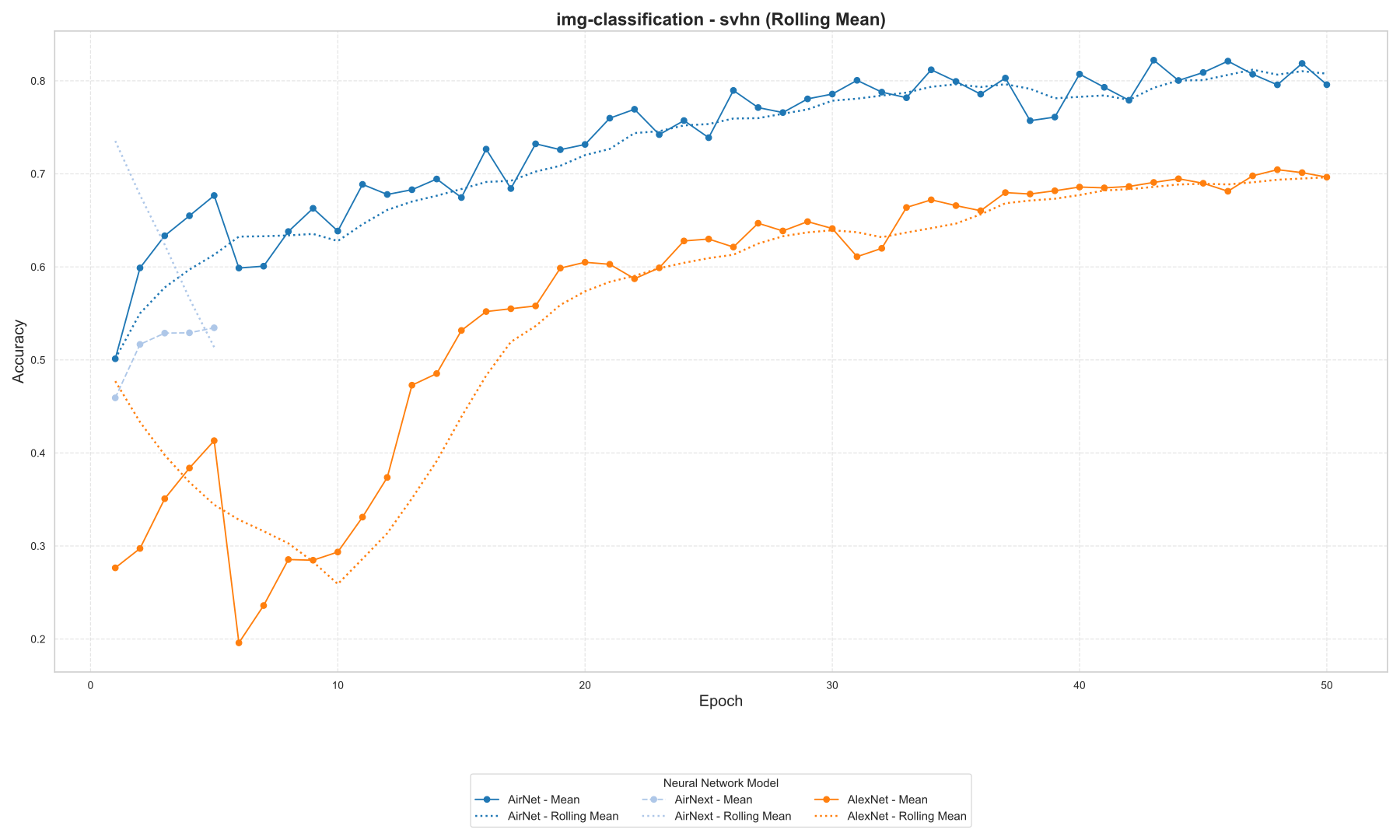}
    \end{minipage}
    \bigskip
    \caption{Rolling Mean Accuracy plots across various datasets. Subfigures (a)--(g) highlight the rolling mean accuracy trends for different classification tasks. Models like AirNet and AlexNet demonstrate varying degrees of efficiency and adaptability across datasets, with notable performance on MNIST, SVHN, and CIFAR datasets. Challenges associated with more complex datasets like Places365 and CIFAR-100 are reflected in the slower improvement trends.}
    \label{fig:rolling_mean_accuracy_grouped}
\end{figure*}

\begin{figure*}[ht]
    \centering
    \begin{minipage}{0.48\linewidth}
        \centering
        \textit{(a)}\\ 
        \includegraphics[width=\linewidth]{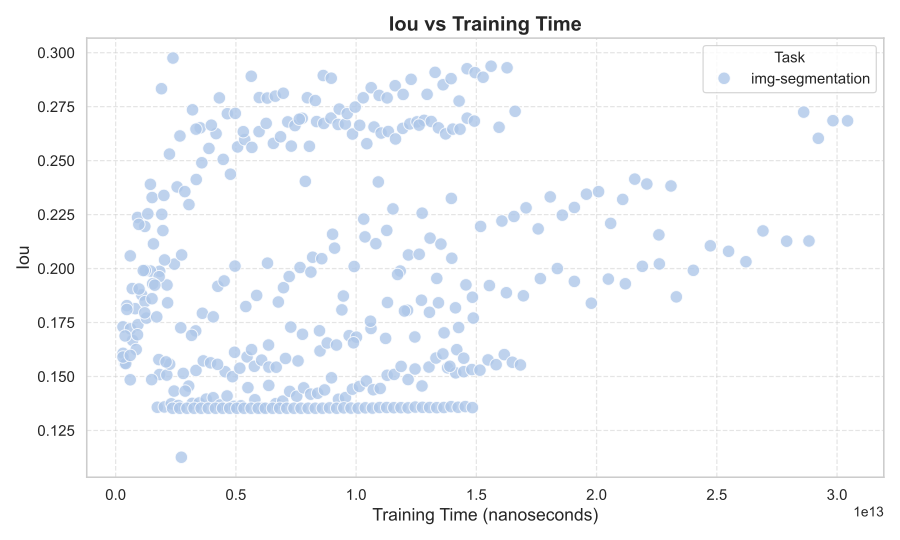}
    \end{minipage}
    \hfill
    \begin{minipage}{0.48\linewidth}
        \centering
        \textit{(b)} \\ 
        \includegraphics[width=\linewidth]{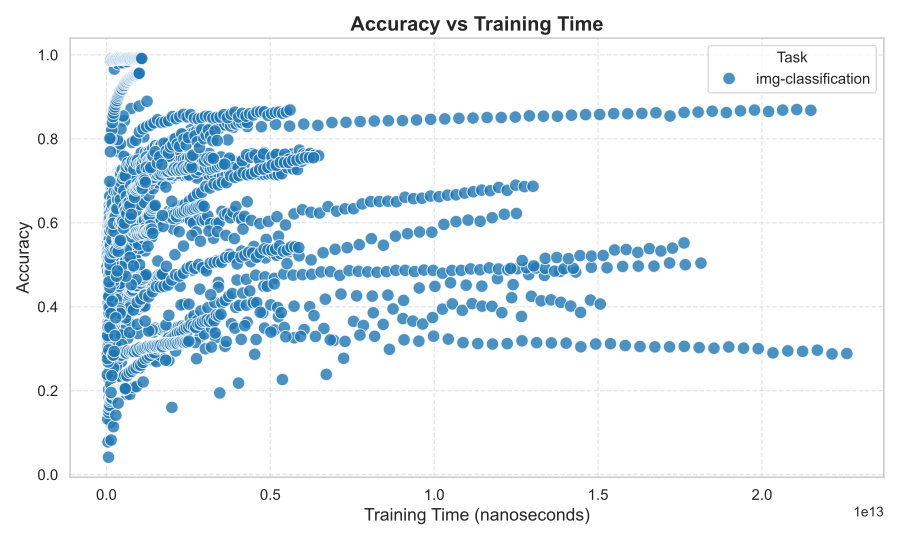}
    \end{minipage}
    \bigskip
    \caption{This figure compares IoU and accuracy against training times for different tasks. Subfigure (a) shows the relationship between IoU scores and training times for image segmentation tasks, highlighting clusters based on segmentation complexity. Subfigure (b) illustrates the correlation between accuracy and training time for image classification tasks, showing the trade-off between accuracy and computational cost.}
    \label{fig:training_time_comparison}
\end{figure*}

\begin{figure}[H]
    \centering
    \includegraphics[width=0.48\textwidth]{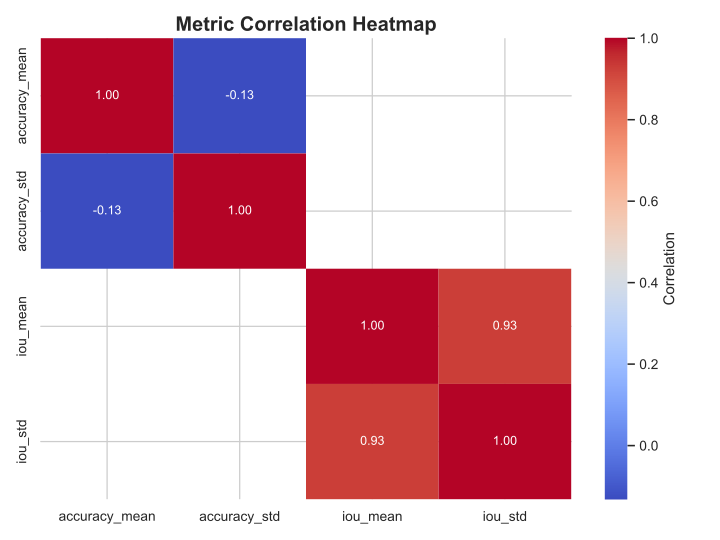}
    \caption{Metric Correlation Heatmap: This heatmap visualizes the correlation coefficients between key metrics such as accuracy mean, accuracy standard deviation (std), IoU mean, and IoU std. A strong positive correlation is observed between IoU mean and IoU std (0.93), indicating consistency in segmentation performance across models. The negative correlation (-0.13) between accuracy mean and accuracy std reflects stability in classification performance as accuracy improves.}
    \label{fig:metric_correlation_heatmap}
\end{figure}

\begin{figure*}[ht]
    \centering
    \begin{minipage}{0.48\linewidth}
        \centering
        \includegraphics[width=\linewidth]{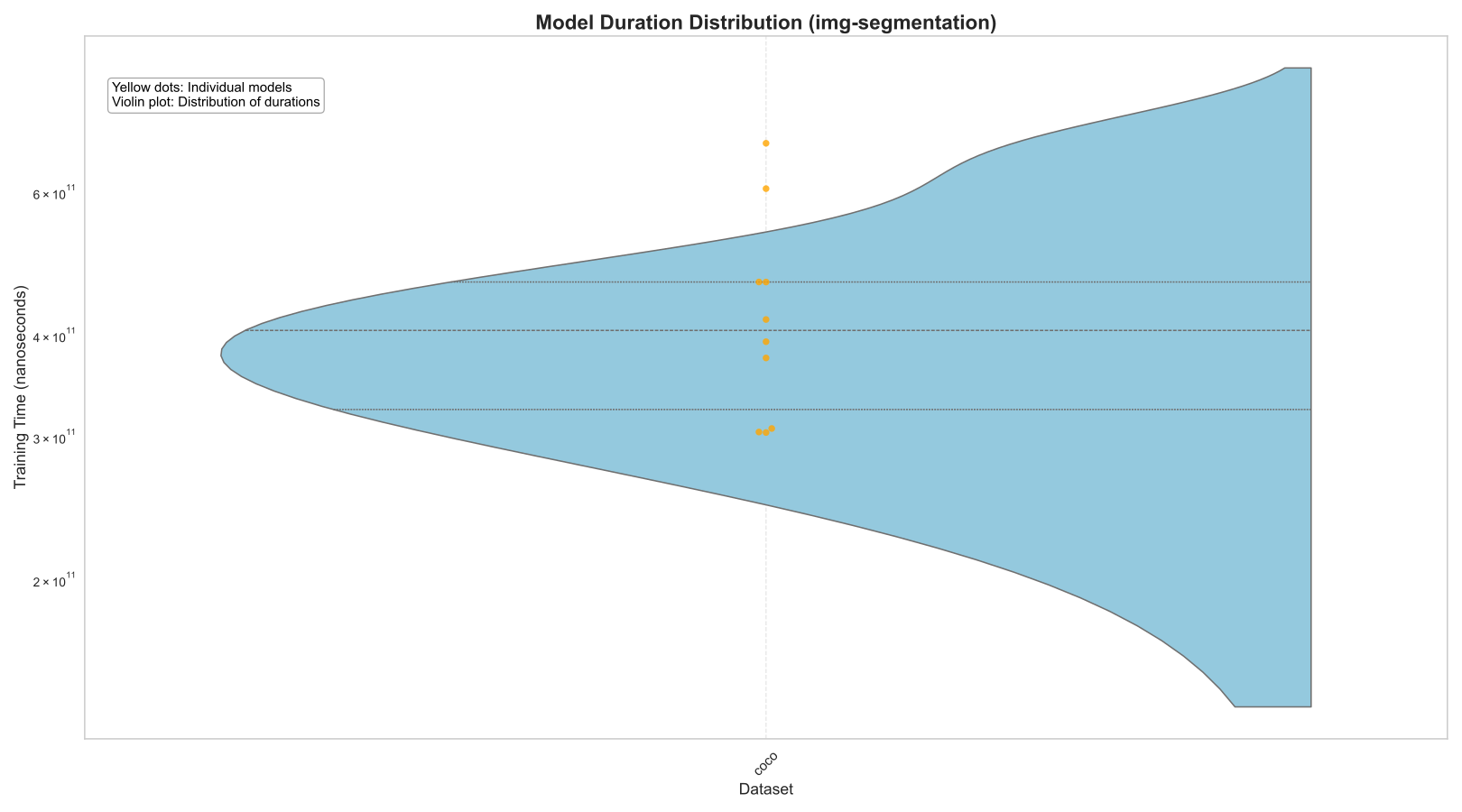}
        \\[5pt]
    \end{minipage}
    \hfill
    \begin{minipage}{0.48\linewidth}
        \centering
        \includegraphics[width=\linewidth]{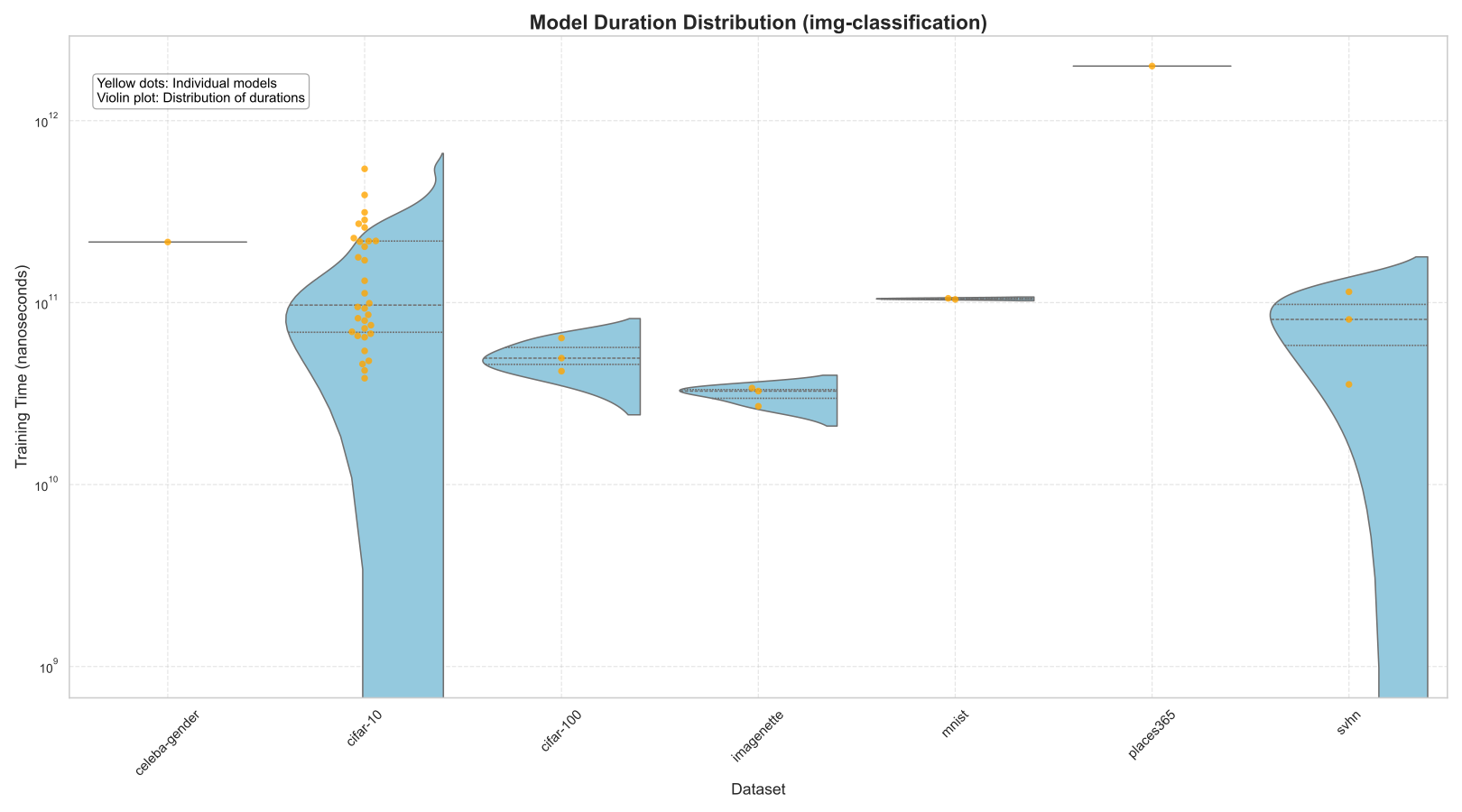}
        \\[5pt]
    \end{minipage}
    \bigskip
    \caption{Distribution of model durations during the first epoch across image segmentation and image classification tasks. The figures highlight computational resource variability for different datasets and models. Image segmentation tasks show clustering around median durations with outliers, while classification tasks reveal variations between datasets.}
    \label{fig:duration_distribution_side_by_side}
\end{figure*}

Line plots (\cref{fig:training_time_comparison}) depict accuracy trends over time for various models and datasets, illustrating performance progression across epochs. Box plots (\cref{fig:accuracy_vs_epochs_box}) and histograms (\cref{fig:accuracy_histogram}) offer statistical perspectives on metric distributions, highlighting variability and common accuracy ranges. Mean and standard deviation plots (\cref{fig:accuracy_trends_with_features}) emphasize trend patterns and variability, shedding light on dataset-specific learning dynamics. Rolling mean plots (\cref{fig:rolling_mean_accuracy_grouped}) smooth out performance noise, revealing clearer convergence behavior. Duration distribution plots (\cref{fig:duration_distribution_side_by_side}) expose computational resource disparities, with image segmentation tasks clustering near median durations and classification tasks showing dataset-dependent variation. Collectively, these visualizations deliver actionable insights into model optimization, dataset complexity, and task-specific challenges—enabling a deeper understanding of neural network behavior.

Box plots (\cref{fig:accuracy_vs_epochs_box}) offer a summary of metric distributions for comparative analysis, highlighting outliers and variability in accuracy across epochs for tasks like image classification, image segmentation, and object detection. For instance, the box plot reveals rapid accuracy stabilization for image classification, contrasted by gradual improvements and wider variability in segmentation and detection tasks. Similarly, histograms (\cref{fig:accuracy_histogram}) provide a statistical perspective, depicting the frequency distribution of accuracy values across tasks. The bimodal distribution seen in \cref{fig:accuracy_histogram} shows significant clustering around 0.2 and 0.6 accuracy, reflecting common performance levels achieved by models across various tasks and datasets. Correlation heatmaps (\cref{fig:metric_correlation_heatmap}) visualize relationships among key metrics, such as accuracy mean, accuracy standard deviation (std), IoU mean, and IoU std. The heatmap highlights a strong positive correlation (0.93) between IoU mean and IoU std, indicating consistent segmentation performance. Conversely, the weak negative correlation (-0.13) between accuracy mean and accuracy std underscores that higher accuracy often corresponds to greater stability in classification models. These insights help researchers understand the interdependence of performance metrics, enabling more targeted model optimization. Scatter plots (\cref{fig:scatter_epochs,fig:scatter_duration,fig:training_time_comparison}) explore trends and interdependencies between training parameters and performance metrics. For instance, the scatter plot of accuracy versus epochs (\cref{fig:scatter_epochs}) illustrates task-specific accuracy trends, where image classification exhibits faster and more consistent improvements compared to image segmentation and object detection. Similarly, the scatter plot of accuracy versus training time (\cref{fig:scatter_duration}) highlights the varying computational demands of tasks: image classification achieves high accuracy with lower training times, while segmentation and detection tasks require significantly longer durations. Additionally, the first-epoch training time distribution (\cref{fig:first_epoch_training_time_distribution_by_model}) emphasizes the computational variability across datasets and models, with simpler datasets like MNIST requiring shorter durations compared to complex datasets like COCO and Places365. Rolling mean plots (\cref{fig:rolling_mean_accuracy_grouped}) provide smoothed representations of fluctuations in performance metrics over time, emphasizing overarching trends. Subplots like CIFAR-10 (\cref{fig:rolling_mean_accuracy_grouped}b) and MNIST (\cref{fig:rolling_mean_accuracy_grouped}e) demonstrate rapid convergence and diminishing variance, while datasets like Places365 (\cref{fig:rolling_mean_accuracy_grouped}f) and CIFAR-100 (\cref{fig:rolling_mean_accuracy_grouped}c) show slower improvements and higher initial variability. These trends highlight the influence of dataset complexity on training dynamics and convergence behaviors.

By integrating these libraries and scripts, the framework ensures that visualizations are both informative and aesthetically refined, empowering researchers to effectively interpret and communicate their results.

\end{document}